%% file: main_elsevier.tex
\documentclass[preprint,11pt,authoryear,times]{elsarticle}
\usepackage[T1]{fontenc} 
\usepackage{nicefrac}
\usepackage[english]{babel} 
\usepackage{framed}
\usepackage{array}

\usepackage{url}
\usepackage{amsmath,amssymb,bm}
\usepackage{graphicx}
\usepackage{lmodern}
\usepackage[caption=false]{subfig}
\usepackage{booktabs,multirow}
\usepackage{arydshln} 
\usepackage[]{units}

\usepackage[font=footnotesize,labelfont=bf,figurename=Fig.]{caption}
\input{LatexDefinitions.tex}

\renewcommand{\cite}[1]{\citep{#1}}
\usepackage{color}
\definecolor{navyblue}{rgb}{0,0,0.5}
\definecolor{winered}{rgb}{0.5,0,0}

\usepackage[normalem]{ulem}
\newcommand{\removed}[1]{\textcolor{navyblue}{\sout{}}}
\newcommand{\rmv}[1]{\textcolor{navyblue}{\sout{}}}

\makeatletter
\renewenvironment{table}{%
  \renewcommand* {\@floatboxreset}{%
    \reset@font\@setminipage}
  \fontsize{8}{10}\@float{table}%
}{%
  \end@float\normalsize
}
\makeatother
\journal{Medical Image Analysis}
\begin{document}

\begin{frontmatter}

\title{Probabilistic Intra-Retinal Layer Segmentation in 3-D~OCT~Images Using Global Shape Regularization}
\author[unihd]{Fabian Rathke\corref{cor1}}
\ead{fabian.rathke@iwr.uni-heidelberg.de}
\author[unihd,hde]{Stefan Schmidt}
\ead{sschmidt@HeidelbergEngineering.com}
\author[unihd]{Christoph Schn\"orr}
\ead{schnoerr@math.uni-heidelberg.de}
\address[unihd]{Image \& Pattern Analysis Group (IPA) and Heidelberg Collaboratory for Image Processing (HCI), University of Heidelberg, Speyerer Str. 6, 69126 Heidelberg, Germany}
\address[hde]{Heidelberg Engineering GmbH, Tiergartenstrasse 15, 69121 Heidelberg, Germany}

\cortext[cor1]{Corresponding author, \textit{Phone}: +49 6221 548787, \textit{Address}: Image \& Pattern Analysis Group (IPA), Speyerer Str. 6, 69126 Heidelberg, Germany, \textit{Email}: fabian.rathke@iwr.uni-heidelberg.de}

\begin{abstract}
With the introduction of spectral-domain optical coherence tomography (OCT), resulting in a significant increase in acquisition speed, 
the fast and accurate segmentation of 3\hbox{-}D OCT scans has become evermore important.
This paper presents a novel probabilistic approach,
that models the appearance of retinal layers as well as the \emph{global} shape variations of layer boundaries.
Given an OCT scan, the \emph{full} posterior distribution over segmentations is approximately inferred using a variational method enabling efficient probabilistic inference in terms of computationally tractable model components: Segmenting a full 3\hbox{-}D volume takes around a minute. Accurate segmentations demonstrate the benefit of using global shape regularization: We segmented 35 fovea-centered 3\hbox{-}D volumes with an average unsigned error of $2.46\pm0.22\,\mu m$ as well as 80 normal and 66 glaucomatous 2\hbox{-}D circular scans with errors of $2.92\pm0.53\,\mu m$ and $4.09\pm0.98\,\mu m$ respectively.  Furthermore, we utilized the inferred posterior distribution to rate the quality of the segmentation, point out potentially erroneous regions and discriminate normal from pathological scans. No pre- or postprocessing was required and we used the same set of parameters for all data sets, underlining the robustness and out-of-the-box nature of our approach. 
\end{abstract}


\end{frontmatter}

\section{Introduction}
\label{chap:introduction}
\input{1_introduction.tex}

\section{Graphical Model}
\input{mod_graph_model}
\section{Variational Inference}
\label{chap:inference}
\input{mod_var_inf}
\section{Optimization}
\label{chap:optimization}
\input{mod_var_opt}

\section{Experiments}
\label{chap:exp-methods}
\input{res_setup}
\subsection{Implementation and Running Time}
\input{res_imp}
\section{Results}
\label{chap:results}
\subsection{Circular Scans}
\input{res_circ}
\subsection{Volumetric Scans}
\label{chap:results-volumes}
\input{res_vol}
\section{Discussion and Conclusion}
\label{chap:discussion}
\input{res_disc}
\section*{Acknowledgments} 

The authors would like to thank Dr.\ Christian Mardin (University hospital Erlangen, Germany) as well as Heidelberg Engineering GmbH for providing the OCT data sets. This work has been supported by the German Research Foundation (DFG) within the program ``Spatio-/Temporal Graphical Models and Applications in Image Analysis'', grant GRK 1653.

\appendix
\input{appendix}

\bibliographystyle{model2-names}
\bibliography{bib/fabian,bib/christoph}
\end{document}

%% file: LatexDefinitions.tex
\newcommand{\bitem}{\begin{itemize}}
\newcommand{\eitem}{\end{itemize}}
\newcommand{\mc}[1]{\mathcal{#1}}
\newcommand{\mb}[1]{\mathbb{#1}}

\newcommand{\R}{\mathbb{R}}

\newcommand{\EE}{\mathbb{E}}
\newcommand{\V}{\mathbb{V}}

\newcommand{\bpm}{\begin{pmatrix}}
\newcommand{\epm}{\end{pmatrix}}
\newcommand{\bsm}{\left(\begin{smallmatrix}}
\newcommand{\esm}{\end{smallmatrix}\right)}

\newcommand{\la}{\langle}
\newcommand{\ra}{\rangle}

\newcommand{\col}[2]{{#1}_{\bullet,#2}}

\newcommand{\sm}[1]{\setminus{#1}}

\newcommand{\mucond}{\mu_{k|k-1,j}}
\newcommand{\sigmacond}{\sigma_{k|k-1,j}^2}

\newcommand{\factor}[1]{A^j_#1}
\newcommand{\factorT}[1]{(A^j_#1)^T}
\newcommand{\factort}[1]{\tilde{A}^j_#1}
\newcommand{\factortT}[1]{(\tilde{A}^j_#1)^T}
\newcommand{\qcjoint}{q_{c}(c_{k,j},c_{k-1,j})}
\newcommand{\qcjointpure}{q_{c;k \wedge k-1,j}}

\DeclareMathOperator{\KL}{KL}

%% file: 1_introduction.tex
\input{intro_motivation}

Various approaches for the task of retina segmentation in OCT images were published.
All have in common that they generate appearance terms based either on intensity or gradient information.
On top of that regularization is applied, which makes predictions more robust to speckle noise or shadowing caused by blood vessels. In order to provide a systematic overview over this vast field of approaches, we choose to distinguish them by the method used for regularization.

One major class is composed of rule-based heuristic techniques~\cite{ahlers2008,fernandez2005,ishikawa2005,mayer2010}, which for example apply outlier detection along with linear interpolation to account for erroneous segmentations. Other approaches \cite{baroni2007,yang2010} use dynamic programming for single Markov chains per boundary and constrain the maximal vertical distance between neighboring boundary positions. \citet{vermeer2011} classify pixels using support vector machines and regularize the output using level-set techniques. None of these approaches incorporates shape prior information.

Active contour approaches include gradient respectively intensity-based methods \cite{mishra2009,yazdanpanah2009,yazdanpanah2011}. 
\citet{yazdanpanah2009,yazdanpanah2011} augment the classical active contour functional by a simple circular shape prior. 
All three approaches were only tested on OCT-scans that exclude the foveal region, thus contain mainly flat boundaries with rather simple shapes. 

A series of more advanced approaches \cite{antony2010,dufour2013,garvin2009,song2013} construct a geometric graph to simultaneously segment all boundaries in a 3-D OCT volume. Unlike previously presented approaches, they take into account the interaction of neighboring boundaries to mutually restrict their relative positions. This shape prior information is encoded into the graph as hard constraints \cite{antony2010,garvin2009} or, as recently introduced by \citet{song2013} and subsequently extended by \citet{dufour2013}, as probabilistic soft constraints. However, due to computational limitations, only \emph{local} shape information is included and boundaries are segmented in stages.


%
Finally, \citet{kajic2010} apply the popular active appearance models that match statistical models for appearance and shape, to a given OCT scan. 
Although non-local shape modeling is in the scope of their approach, they only use landmarks, i.e. sparsely sampled boundary positions instead of the full shape model.
Furthermore, only a maximum likelihood point estimate is inferred, instead of a distribution over shapes.
\\[0.3cm]
\textbf{Contribution.} We present a novel probabilistic approach for the OCT retina segmentation problem. Our probabilistic graphical model combines appearance models with a global shape prior, that comprises local as well as long-range interactions between boundaries. The discrete part of the model features a highly parallelizable column-wise discrete segmentation, that nevertheless takes into account all other image columns. In order to infer the posterior probability of this model, we utilize variational inference, a deterministic approximation framework. 

\begin{figure}
\centering
\includegraphics[width=1\textwidth]{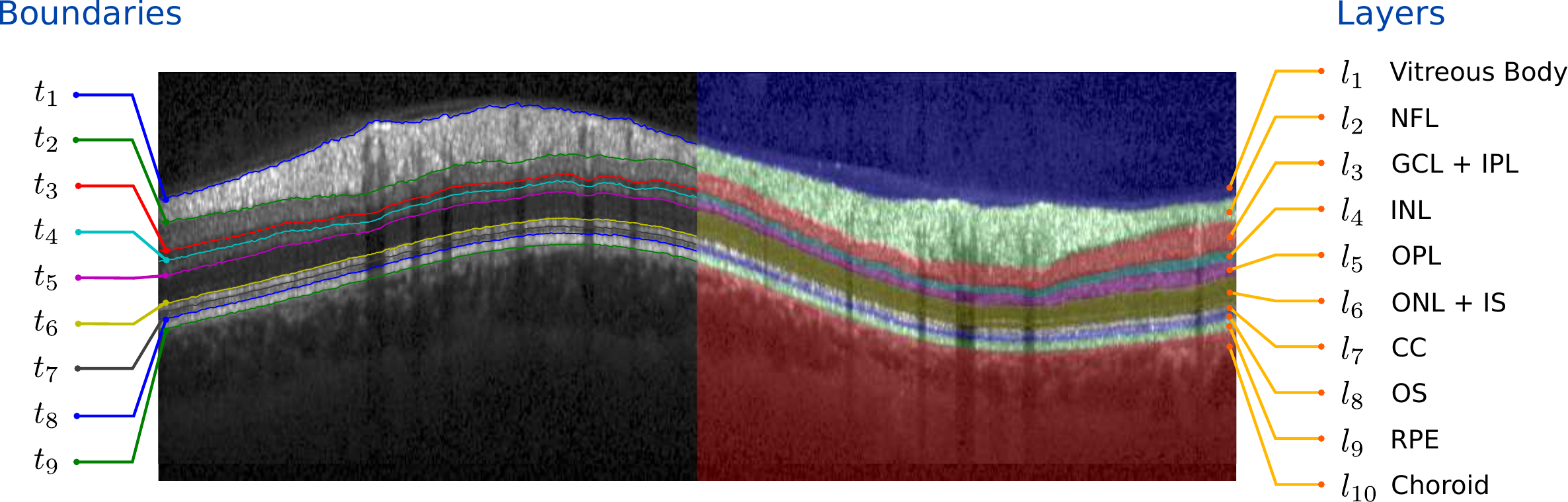}
\caption{Overview of retinal layers segmented by our approach and their corresponding anatomical names. The used abbreviations correspond to nerve fibre layer (NFL), ganglion cell layer and inner plexiform layer (GCL~+~IPL), inner nuclear layer (INL), outer plexiform layer (OPL), outer nuclear layer and inner segment (ONL + IS), connecting cilia (CC), outer segment (OS), retinal pigment epithelium (RPE).}
\label{fig:overview_segmented_layers}
\end{figure}

To our knowledge this is the only work,
where a full \emph{global} shape prior is employed for the task of OCT retina segmentation. Moreover, we are not aware of any other segmentation approach that infers a full probability distribution. Our approach offers excellent segmentation performance, outperforming approaches relying on local or no shape regularization,
as well as pathology detection and an assessment of segmentation quality.
Fig.~\ref{fig:overview_segmented_layers} illustrates the segmented boundaries, but additional boundaries like the external limiting membrane (ELM) could easily be incorporated if ground truth is available. 

This work evolved out of preliminary ideas presented in a previous conference paper \cite{rathke2011b}. 
\\[0.3cm]
\textbf{Organization.} The next section will introduce our probabilistic graphical model. Section \ref{chap:inference} 
evaluates the posterior distribution via variational inference, and we solve the corresponding optimization problem in Section~\ref{chap:optimization} in terms of efficiently solvable convex subproblems. Section \ref{chap:exp-methods} and \ref{chap:results} present the data sets we used for evaluation and the corresponding results. We conclude in Section~\ref{chap:discussion} with a discussion and possible directions for future work.

%% file: intro_motivation.tex
Optical coherence tomography (OCT) is an \textit{in vivo} imaging technique, measuring the delay and magnitude of backscattered light. Providing micrometer resolution and millimeter penetration depth into retinal tissue \cite{drexler2008}, OCT is well suited for ophthalmic imaging. Since no other method can perform noninvasive imaging with such a resolution, OCT has become a standard in clinical ophthalmology \cite{schuman2004}. Several studies showed the applicability for the diagnosis of pathologies such as glaucoma or age-related macular degeneration \cite{bowd2001,zysk2007}. The recent introduction \cite{deboer2003,wojtkowski2002} of spectral-domain OCT dramatically increased the imaging speed and enabled the acquisition of 3\hbox{-}D volumes containing hundreds of B-scans. Since manual segmentation of retinal layers is tedious and time-consuming, automated segmentation becomes evermore important given the growing amount of gathered data. Furthermore, a probabilistic model that enables to infer uncertainties of estimates, provides essential information for practitioners, in addition to the segmentation result. 

%% file: mod_graph_model.tex
\begin{figure}
\centering
\begin{minipage}[t]{.44\textwidth}
  	\centering
	\includegraphics[width=1\textwidth]{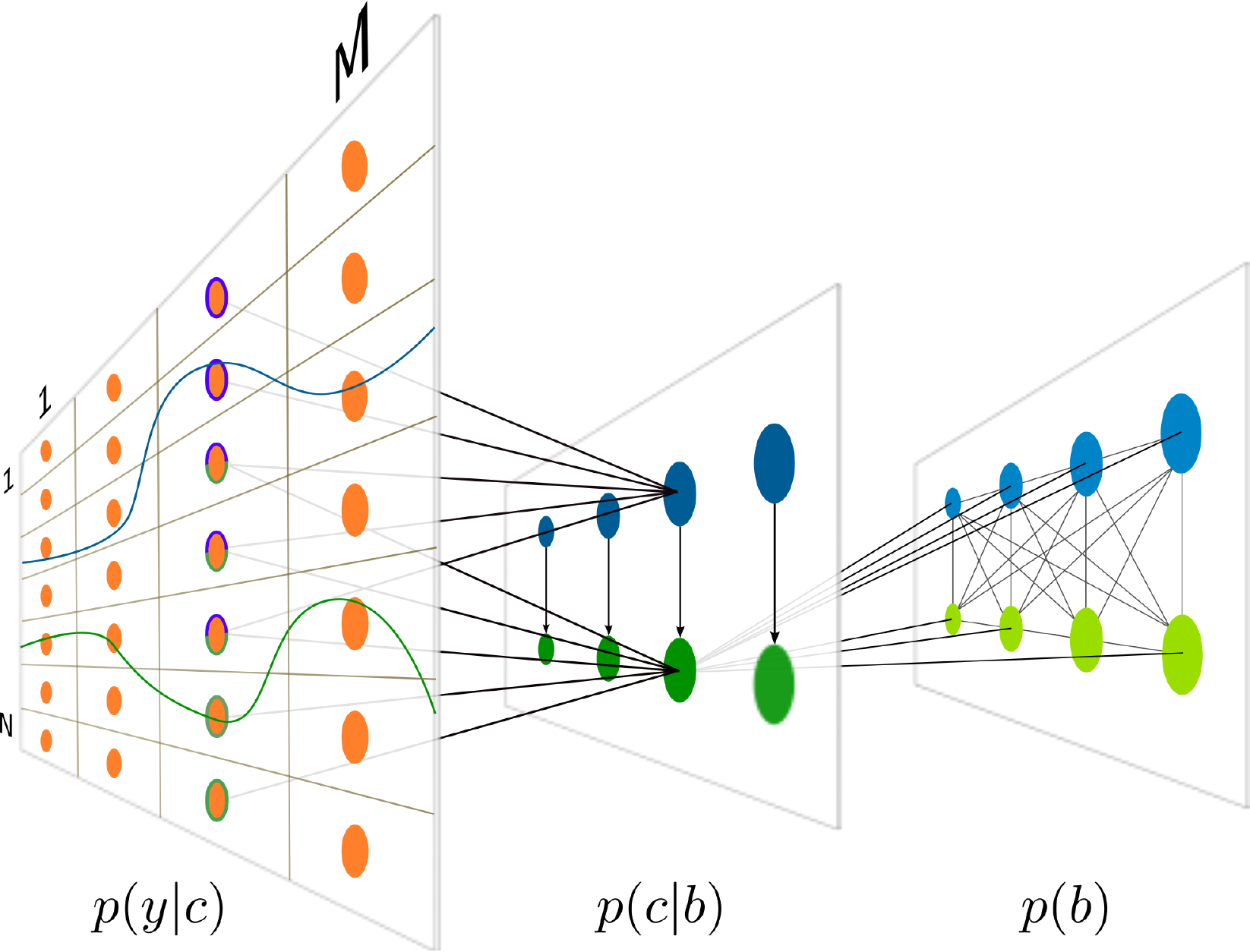}
	\caption{Illustration of our graphical model for $M = 4$, $N = 7$ and $N_b = 2$. The connectivity from $b$ to $c$ is only displayed for node $c_{2,3}$. Similarly, connectivity for $c$ to $y$ via $x$ is only displayed for nodes in the third image column and additionally illustrated by the boundary colors of the $y$-nodes.}
	\label{fig:graph_model}
\end{minipage}
\hfill
\begin{minipage}[t]{.52\textwidth}
  \centering
	\includegraphics[width=1\textwidth]{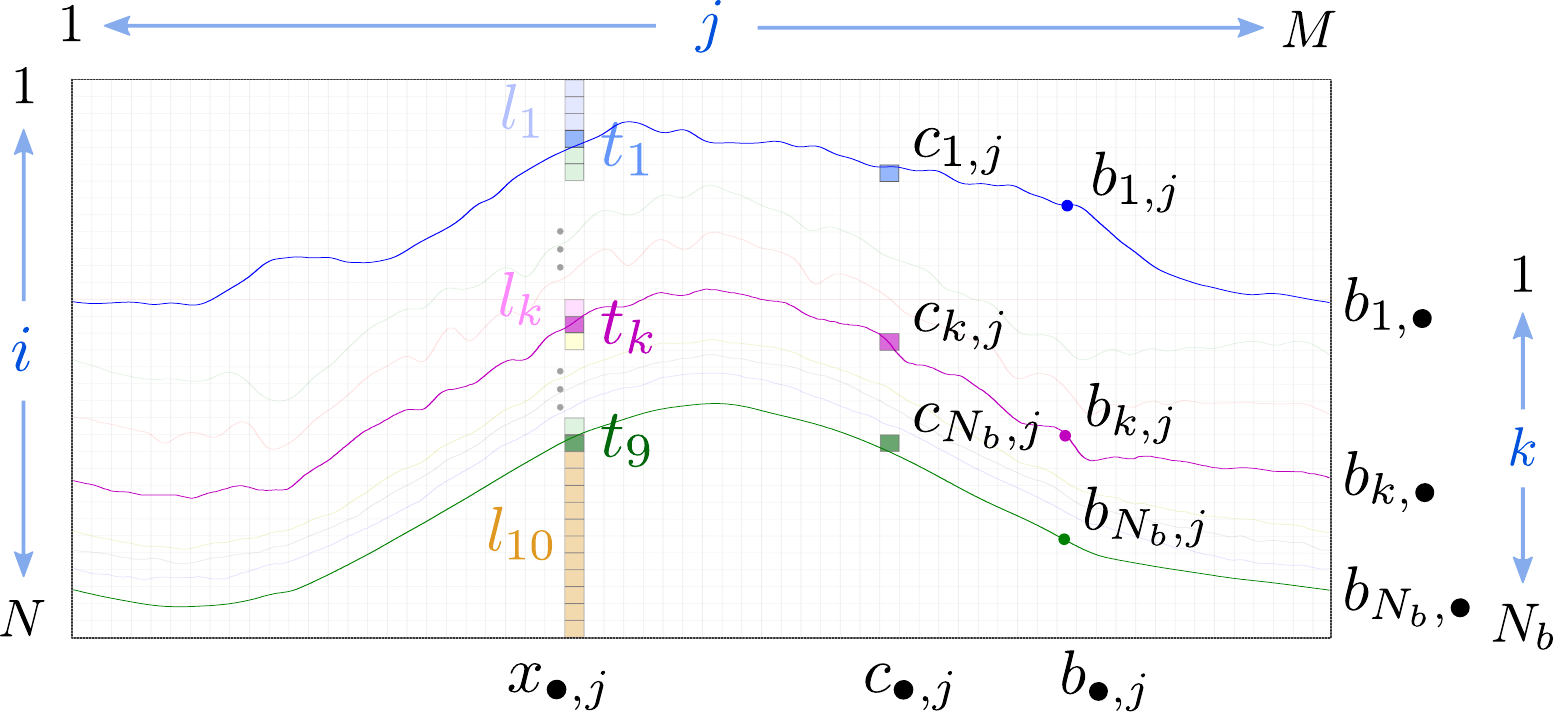}
	\caption{Illustration of important variables used throughout the paper. Note the difference between real valued position $b_{k,j}$ of boundary $k$ in column $j$ 	and its discretized equivalent $c_{k,j}$.}
\label{fig:notation}
\end{minipage}
\end{figure}

This section presents our probabilistic graphical model, statistically modeling an OCT scan $y$ and its segmentations $b$ and $c$ respectively. We introduce $c$, the discretized version of the continuous boundary vector $b$, to make mathematically explicit the connection between the discrete pixel domain of $y$ and the continuous boundary domain of $b$. Our ansatz is given by
\begin{equation}
p(y,c,b) = p(y|c) p(c|b) p(b),
\label{eq:ansatz-model}
\end{equation}
where the factors are \\[0.2cm]
\begin{tabular}{ll}
$p(y|c)$ & appearance, data likelihood term, \\
$p(c|b)$ & Markov Random Field regularizer, determined by the shape prior and \\
$p(b)$ & global shape prior.
\end{tabular} 
\\[0.3cm]In what follows we will detail each component, thereby completing the definition of our graphical model. Fig. \ref{fig:graph_model} illustrates our graphical model in terms of the connectivity of the individual model layers.
\input{notation}

\subsection{Appearance Models}
We utilize Gaussian distributions to model the appearance of retinal layers as well as boundaries. Given a segmentation hypothesis $c$, we can assign class labels $x_{i,j} \in \mc{X}$ to each pixel, their range being given by 
\begin{equation*}
\mc{X} = \{\mc{X}_l,\mc{X}_t\}, \qquad \mc{X}_l = \{l_{1},\dotsc,l_{10}\}, \quad \mc{X}_t = \{t_{1},\dotsc,t_{9}\},
\end{equation*}
which represent classes of observations corresponding to tissue layers $l_{1},\dotsc,l_{10}$ and transitions $t_{1},\dotsc,t_{9}$ separating them. To obtain a valid mapping $c \mapsto x$, we require $c$ to satisfy the ordering constraint \removed{in agreement with physiology},
\begin{equation} \label{eq:c-ordering}
 1 \leq c_{1,j} < c_{2,j} < \dotsb < c_{N_{b},j} \leq N,\quad
 \forall j = 1,\dotsc,M,
\end{equation}
and point out that the real-valued counterpart $b$ may violate this constraint.

Since OCT scans display a high variability in brightness and contrast within and between scans, each patch is first normalized by subtracting its mean. We then project each patch $y_{i,j}$ onto a low-dimensional manifold. Applying the technique of Principle Component Analysis (PCA), we draw patches randomly from the OCT scans in the training set, estimate their empirical covariance matrix and calculate its eigenvalues and eigenvectors. The projection can then be carried out using the first $q_{\mathrm{pca}}$ eigenvectors sorted by their eigenvalues.

We define the probability of the projected patch $y_{i,j}$\footnote{For ease of notation, we will make no difference between the patch $y_{i,j}$ and its low-dimensional projection.} around pixel $(i,j)$ belonging to class $x_{i,j}$ as
\begin{equation}
\label{eq:generative-app-terms}
p(y_{i,j} | x_{i,j}(c)) = \mc{N}(y_{i,j};\mu_{x_{i,j}},\Sigma_{x_{i,j}})\,.
\end{equation}
The class-specific moments $\mu_x, \Sigma_x \; \forall x \in \mc{X}$ are learned offline using patches from the respective class. Regularized estimates for $\Sigma_x$ are obtained by utilizing the graphical lasso approach \cite{friedman2008}, which augments the classical maximum likelihood estimate for $\Sigma$ with an $\ell_1$-norm on the precision matrix $K = \Sigma^{-1}$. This leads to sparse estimates for $K$, where the degree of sparsity is governed by a parameter $\alpha_{\mathrm{glasso}}$, c.f. Section \ref{chap:parameters}.

We define the appearance of a scan $y$ to factorize over pixels $(i,j)$.
Finally, we introduce switches $\beta^{t} \in \{0,1\}$ and $\beta^{l} \in \{0,1\}$, that turn on/off all terms belonging to any transition class $t_k$ or layer class $l_k$, which yields the final appearance model
\begin{equation}
\label{eq:appearance-terms}
p(y|c) = \prod_{j=1}^M \prod_{i:x_{i,j} \in \mc{X}_l} \hspace{-0.2cm} p(y_{i,j}|x_{i,j}(c))^{\beta^{l}} \hspace{-0.3cm} \prod_{i:x_{i,j} \in \mc{X}_t} \hspace{-0.2cm} p(y_{i,j}|x_{i,j}(c))^{\beta^{t}}.
\end{equation}
As we point out in the section about inference, our model can handle discriminative terms as well. We can convert generative terms~\eqref{eq:generative-app-terms} into discriminative ones by renormalizing 
\begin{equation}
\label{eq:discriminative-app-terms}
p(x_{i,j}(c)|y_{i,j}) = \frac{p(y_{i,j}|x_{i,j}(c))p(x_{i,j}(c))}{\sum_{x_{i,j} \in \mc{X}} p(y_{i,j}|x_{i,j}(c))p(x_{i,j}(c))}, 
\end{equation}
where we use a uniform prior $p(x_{i,j}(c))$. The factorization of $p(c|y)$ is the same as in \eqref{eq:appearance-terms}.

\subsection{Shape Prior}
\label{sec:shape-prior}
As a model of the typical shape variation of layers due to both biological variability as well as to the image formation process, we adopt a joint Gaussian distribution\footnote{For circular scans, a wave-like distortion pattern is observed due to the conic scanning geometry and the spherical shape of the retina, which we capture statistically rather than modeling it explicitly.}. We denote the continuous height values of all boundaries $k$ for image columns $j$ by the $N_bM$-dimensional vector $b = (b_{k,j})_{\substack{k=1,\dots,N_b;\, j=1,\dots,M}}$. Hence,
\begin{equation} \label{eq:shape-prior}
p(b) = \mc{N}(b;\mu,\Sigma) \,
\end{equation}
where parameters $\mu$ and $\Sigma$ are learned offline from labeled training data. We regularize the estimation of $\Sigma$ by Probabilistic Principal Component Analysis (PPCA) \cite{tipping1999}. PPCA assumes that the high-dimensional observation $b$ was generated from a low-dimensional latent source $s \in \R^q$ via
\begin{equation*}
b = Ws + \mu + \epsilon \,,
\end{equation*}
where $s \sim \mc{N}(0,I)$ and $\epsilon \sim \mc{N}(0,\sigma^2I)$ is isotropic Gaussian noise\footnote{PPCA can be considered as a generalisation of classical Principle Component Analysis (PCA), which assumes the deterministic relation $b = Ws + \mu$.}.

The moments of $p(b)$ are given by $\mb{E}[b] = \mu$ and $\mb{V}[b] = WW^T + \sigma^2I = \Sigma$. Likewise, $\Sigma^{-1}$ can be decomposed into $W$ and $\sigma^2I$ too. Making use of these decompositions, one can reduce complexity of most operations related to $\Sigma$ or $\Sigma^{-1}$ as well as memory requirements, since only $W$ and $\sigma^2$ have to be stored. The parameters for $p(b)$ can be estimated via maximum log-likelihood. $W$ is composed of the (weighted) $q_{\mathrm{ppca}}$~eigenvectors with largest eigenvalues, computed from the empirical covariance matrix. For more details, we refer to \citet{tipping1999}.

\begin{figure}[t]
\centerline{
\subfloat{\includegraphics[width=0.47\textwidth]{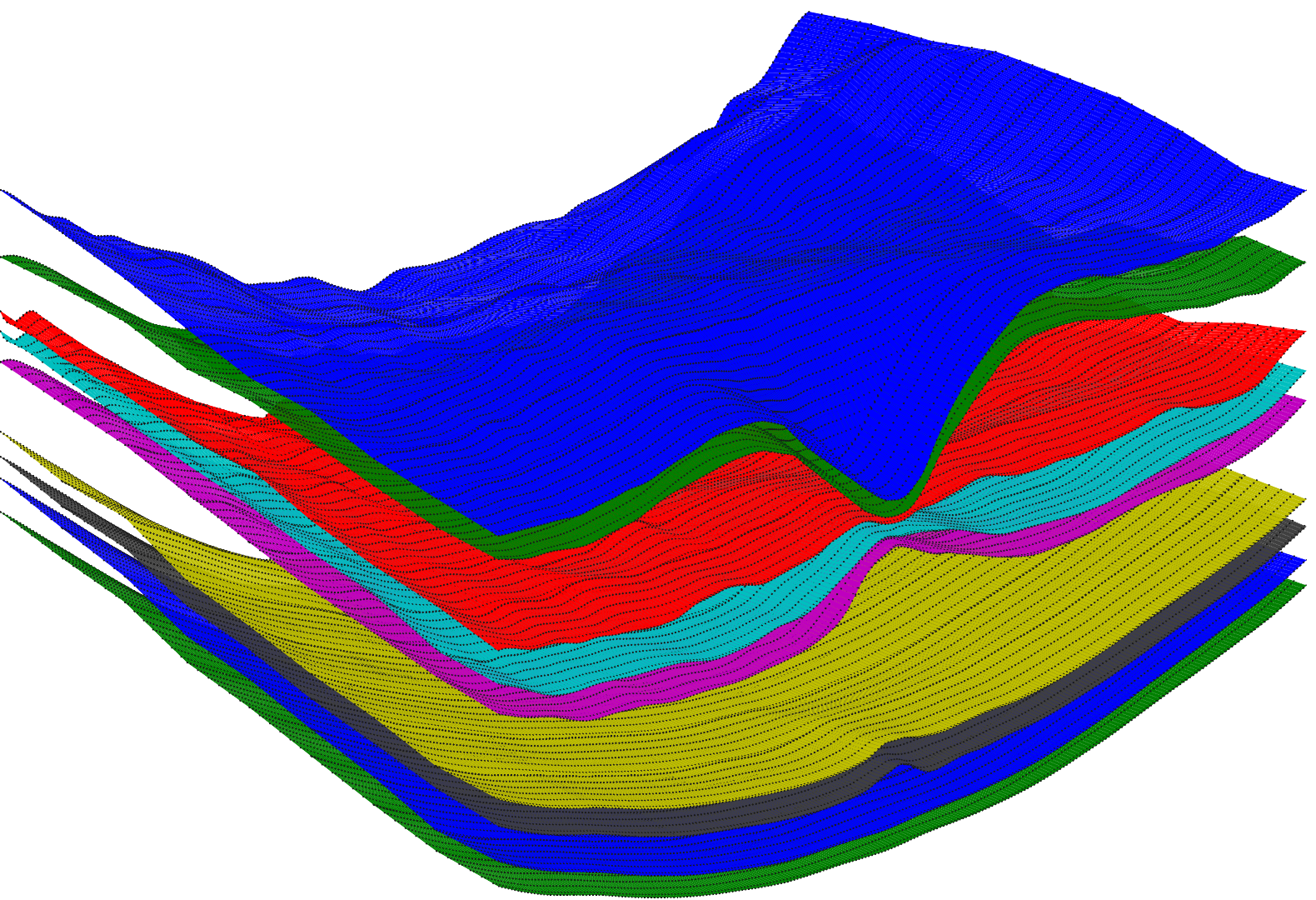}}\hfill
\subfloat{\includegraphics[width=0.47\textwidth]{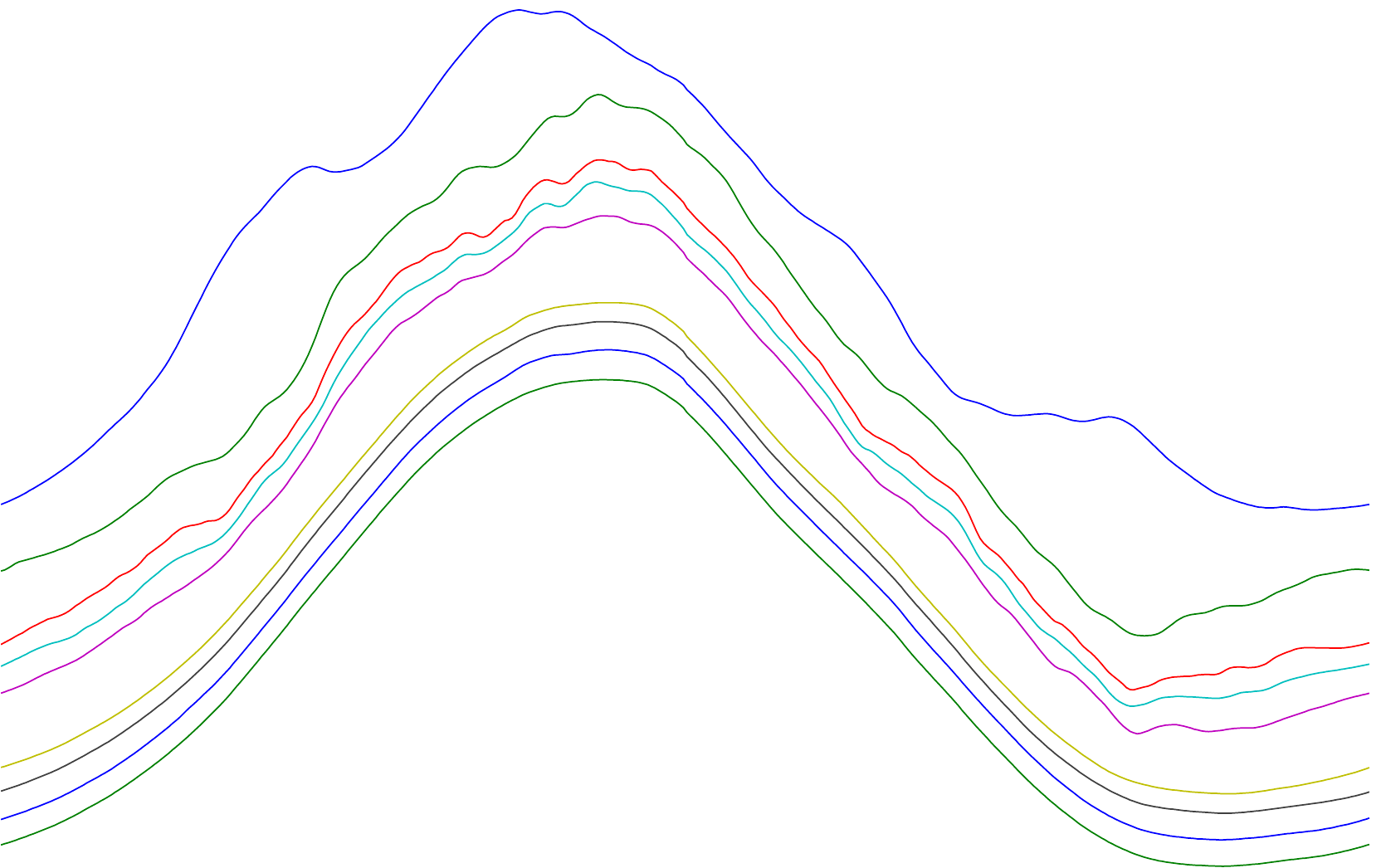}}
}
\vspace{0.5cm}
\centerline{
\subfloat{\includegraphics[width=0.47\textwidth]{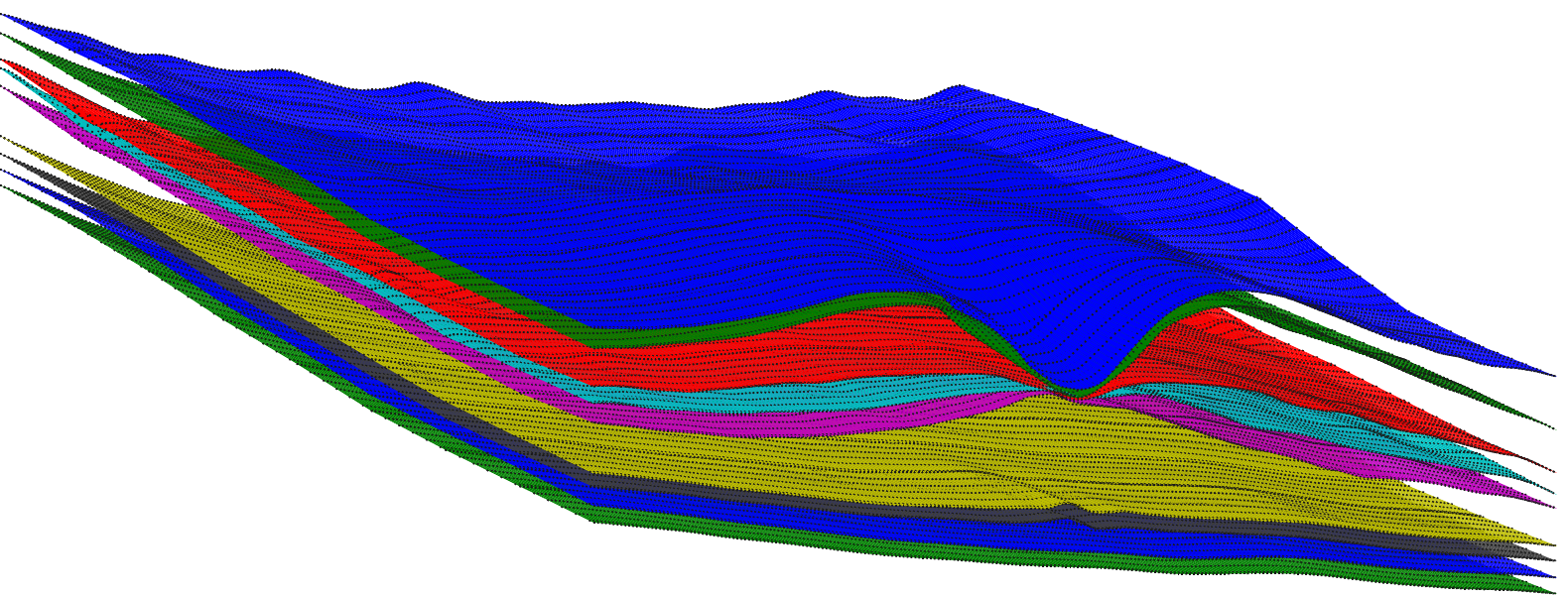}}\hfill
\subfloat{\includegraphics[width=0.47\textwidth]{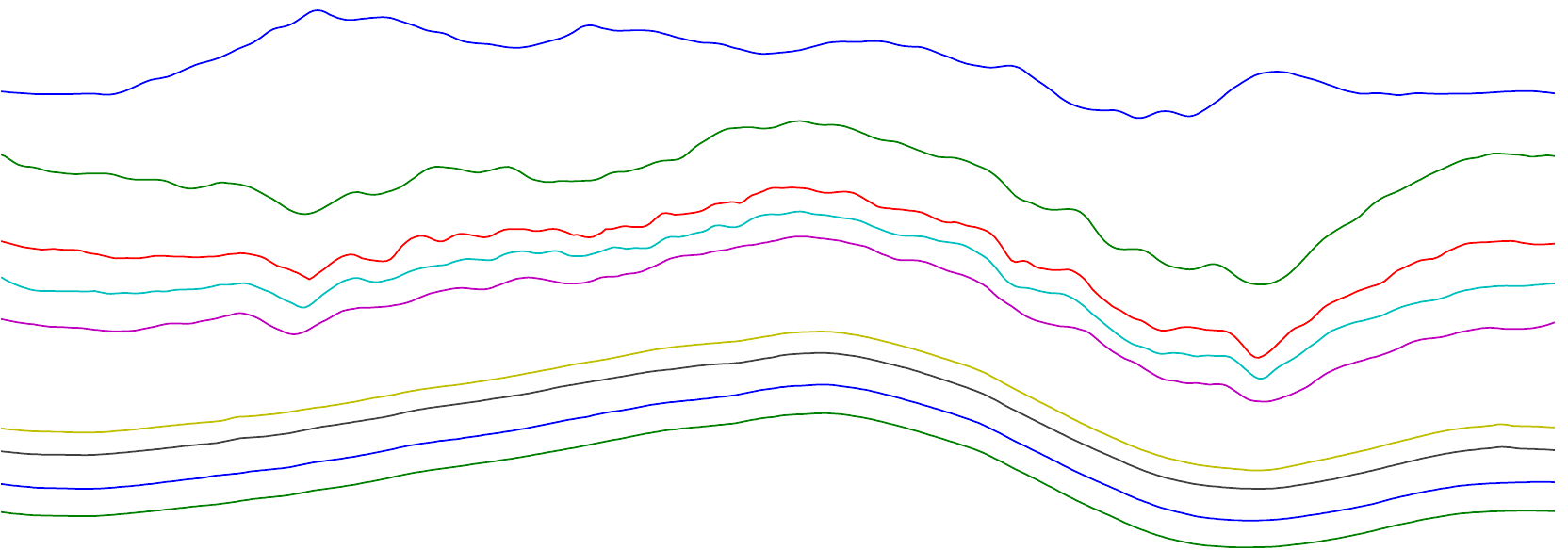}}
}
\caption{Samples generated by the shape prior distribution $p(b)$ trained on volumes (left) and circular scans (right). Only one half of the volume is shown.}
\label{fig:samples}
\end{figure}

Fig. \ref{fig:samples} shows samples drawn from $p(b)$, modeling fovea-centered 3-D volumes~(left) and circular scans~(right). Additionally, as supplementary material we provide a video that visualizes the relevance of each component of $W$ exemplary for the 3-D setup, like translation, rotation, thickness of layers or position and form of the fovea.

\subsection{Shape-Induced Regularizers}
The third component of our model consists of a prior for discrete boundary assignments~$c$, regularizing the data likelihood terms~$p(y|c)$. We define~$p(c|b)$ as column-wise acyclic graphs
\begin{equation}
\label{eq:regularizer-basic}
p(c|b) = \prod_{j=1}^{M} p(\col{c}{j}|b),\qquad
p(\col{c}{j}|b) = p(c_{1,j}|b) \prod_{k=2}^{N_b} p(c_{k,j}|c_{k-1,j},b),
\end{equation}
i.e. the communication of the model \textit{between} image columns $j$ is governed by the shape~prior~$p(b)$.

In order to define the conditional marginals in \eqref{eq:regularizer-basic}, we need a couple of prerequisites. With $\col{b}{\setminus j}$ denoting the subset of variables $b$ after removing variables $\col{b}{j}$ of column $j$, and with $p(\col{b}{j}|\col{b}{\setminus j})$ denoting the corresponding conditional Gaussian distribution computed from the shape prior $p(b)$, then the marginal distributions are specified in terms of $b$ by
\begin{equation}
\begin{aligned}
 p(c_{1,j}\!=\!n|b) &= \Pr\Big(n\!-\!\frac{1}{2}\!\leq b_{1,j}\!\leq n\!+\!\frac{1}{2}\,\Big), \\
 p(c_{k,j}\!=\!n|c_{k-1,j}\!=\!m,b) &= \\
   \Pr\Big(n\!-\!\frac{1}{2}\!\leq b_{k,j}&\!\leq n\!+\!\frac{1}{2} \,\Big|\, m\!-\!\frac{1}{2} \leq b_{k-1,j}\!\leq m\!+\!\frac{1}{2}\,\Big),
\end{aligned}
\label{eq:c-given-b}
\end{equation}
where the probabilities on the right-hand side ~are computed using the conditional marginal densities $p(b_{1,j}|\col{b}{\setminus j})$ and $p(b_{k,j}|\col{b}{\setminus j})p(b_{k,j}|b_{k-1,j})$ respectively, for all configurations of $c$ conforming to~\eqref{eq:c-ordering}\footnote{This computation is straightforward for Gaussian distributions \eqref{eq:shape-prior}, see Section \ref{chap:p(c|b)}.}. The marginal $p(b_{k,j}|\col{b}{\setminus j})$ provides a way to introduce global shape knowledge into our column-wise graphical models $p(c|b)$. 


\input{mod_extension}

%% file: notation.tex
\\[0.3cm]
\textbf{Notation.} The following notation is used throughout the paper:

\vspace{0.4cm}\begin{tabular}{>{\small}l>{\small}l}
$N,M$
&OCT scan dimensions (rows, columns); \\
$N_b$
& number of segmented boundaries; $N_b = 9$ in this paper;\\
$i,j,k$
& corresponding indices: \\
& $i=1,\dotsc,N,\; j=1,\dotsc,M, \; k=1,\dotsc,N_b$; \\
$b_{k,j} \in \R$
& real-valued location of boundary $k$ in column $j$; \\
$c_{k,j} \in \{1,\dotsc,N\}$
& integer-valued boundary variables analogous to $b$, \\
& but specifying row-positions on the pixel grid;\\
$x_{i,j} \in \mc{X}$
& class variables indicating membership to \\ 
& layer or transition classes;\\
$y_{i,j}$ 
& observed data; here patches around pixel $(i,j)$\\
& projected onto a low-dimensional manifold \\
$\Delta_{N}$ 
& standard ($N\text{-}1$)-simplex: $\forall \theta \in \Delta_{N}: \sum_{i=1}^N \theta_i = 1$ \\
\end{tabular}
\vspace{0.4cm} \\
The symbol $\bullet$ denotes the set of all elements of the respective index, for example $b_{k,\bullet} \in \R^{M}$ is the location vector for boundary $k$. By $b_{\setminus{j}}$ we denote the set $b \setminus b_{\bullet,j}$, with similar notations used for $\mu$ and $\Sigma$. See Fig. \ref{fig:notation} for an illustration of most of the notation introduced here.

%% file: mod_extension.tex
\subsection{2-D vs. 3-D}
Our description so far considered OCT scans $y$ of dimensionality two. Nevertheless, our approach is equally applicable to 3-D volumes. We use the very same notation, since adding additional B-Scans will only increase the number of image columns $M$. Similarly, the connectivity of the graphical model $p(y,c,b)$ can be transferred one-to-one.

The shape prior $p(b)$ which is fully connected since $\Sigma^{-1}$ is dense, can be extended to an arbitrary dimension. We exploit the fact that both, $\Sigma$ and $\Sigma^{-1}$, have an explicit low-rank decomposition (see section \ref{sec:shape-prior}), such that memory consumption is not an issue and complexity of operations is reduced as well. For the shape regularization term $p(c|b)$, each node $c_{k,j}$ is connected to nodes $b_{\sm j}$ of all columns except the current one, which now additionally includes columns of all other B-scans. Finally, the data likelihood $p(y|c)$ continues to fully factorize over pixels $(i,j)$. Each pixel $(i,j)$ remains connected to at most two nodes $c_{k,j}$ from the same column~$j$, determining it's label $x_{i,j}$. Furthermore, we use separate sets of appearance models for each B-scan in the volume to capture possible variations.


%% file: mod_var_inf.tex
Based on the model presented in the previous section and given observed data $y$, we wish to infer the posterior
\begin{equation}
p(b,c|y) = \frac{p(y|c) p(c|b) p(b)}{p(y)}.
\end{equation}
One major issue is that calculating the marginal likelihood $p(y)$ would require integrating $p(y,c,b)$ over $b$~and~$c$. This turns out to be intractable, since we lack a closed form solution and the problem at hand is high-dimensional. We cope with this problem by applying an established variational method: approximating the posterior by a tractable distribution $q(b,c)$ by minimizing the Kullback-Leibler (KL) distance $\KL(q\|p)$ with respect to $q$ (cf., e.g., \citet{VariationalBayes-99}). We point out that unlike in related work (e.g.~\citet{VariationalBayesHiddenPotts-09}) where the subproblem of inferring the discrete decision variables has to be approximated as well, our model has been designed such that by choosing $q$ properly all subproblems are tractable and can be solved efficiently. 

We choose a factorized approximating distribution 
\begin{equation}
q(b,c) = q_{b}(b) q_{c}(c).
\end{equation}
This merely decouples the continuous shape prior and the discrete order-preserving segmentation component of the overall model, but otherwise represents both components exactly. The Kullback-Leibler distance between $q$ and $p$ is given by
\begin{align}
\nonumber \KL\big(&q(b,c)\big\|p(b,c|y)\big) = 
\int_{b} \sum_{c} q(b,c) \log\frac{q(b,c)}{p(b,c|y)} db \\
&= -\int_{b} \sum_{c} q(b,c) \Big(
\log\big(p(y|c) p(c|b) p(b)\big) - \log p(y) - \log q(b,c) \Big) db \,.
\label{eq:kl_minimization}
\end{align}
Dropping the constant term $\log p(y)$, we may obtain our objective function. 

Alternatively, we can use the marginal likelihood $\log p(y)$ to introduce \emph{discriminative} appearance terms into the model, using $\log\frac{p(y|c)}{p(y)} = \log\frac{p(y|c) p(c)}{p(y)} - \log p(c) = \log p(c|y) - \log p(c)$.  Since $p(b)$ already contains prior knowledge about the shape of boundary positions, we assume an uninformative prior for $c$. Dropping thus $p(c)$ and taking into account the factorization of $q$, we obtain the objective function
\begin{equation} \label{eq:J-functional}
\begin{aligned}
 J(q_{b},q_{c}) = -\int_{b} \sum_{c} q_{b}(b) q_{c}(c)
 \log\big(p(c|y) p(c|b) p(b)\big) db - H_{q_b}(b)  - H_{q_c}(c),
\end{aligned}
\end{equation}
where $H_p(x)$ denotes the entropy of the distribution $p$. It turned out that discriminative appearance terms yielded the best performance. A discussion of this issue 
will be given in Section 5.2.
\\[0.3cm]
\textbf{Definitions of $q_c$ and $q_b$.} For $q_c(c)$, we adopt the same factorization as for $p(c|b)$, that is
\begin{equation}
\label{eq:q-factorized}
q_{c}(c) = \prod_{j=1}^M q_{c;1,j}(c_{1,j}) \prod_{k=2}^{N_b} \frac{q_{c;k\wedge k-1,j}(c_{k,j},c_{k-1,j})}{q_{c;k-1,j}(c_{k-1,j})}
\end{equation}
where $q_{c;k,j} \in \Delta_{N}$ are discrete probability distributions. Similarly, by $\qcjointpure \in \Delta_{N^2}$ we denote discrete probability distributions over pairs of variables $c_{k-1,j},c_{k,j}$\footnote{To enhance readability, we will subsequently omit indices $k,j$ of $q_c$, if they are determined by the input variable(s) $c_{k,j}$.}. For $q_c(c)$ to be a valid distribution, additional marginalization constraints have to be satisfied:
\begin{equation}
\label{eq:q-c-pairwise-constraints}
\sum_{c_{k-1,j}} \qcjoint = q_{c}(c_{k,j}), \quad \sum_{c_{k,j}} \qcjoint = q_c(c_{k-1,j})\,,
\end{equation}
for all $k=2,\ldots,N_b$ and $j = 1,\ldots,M$. Note that we ignore the set of valid configurations \eqref{eq:c-ordering} here, because this has already been taken into account when defining $p(c|b)$. As for $q_c$ and $p(c|b)$, we let $q_b$ adopt the same factorization as $p(b)$, thus 
\begin{equation}
\label{eq:q-b-def}
q_b(b) = \mathcal{N}(b; \bar{\mu},\bar{\Sigma})\,.
\end{equation}

In what follows, we make the expectations with respect to $q_c$ and $q_b$ explicit. This will provide us below with a closed-form expression of the objective function $J(q_b,q_c)$. 


\subsection{First Summand $\log p(c|y)$ of $J(q_b,q_c)$}
\label{chap:p(c|y)}
\begin{figure}[!t]
\centering
\includegraphics[width=0.9\textwidth]{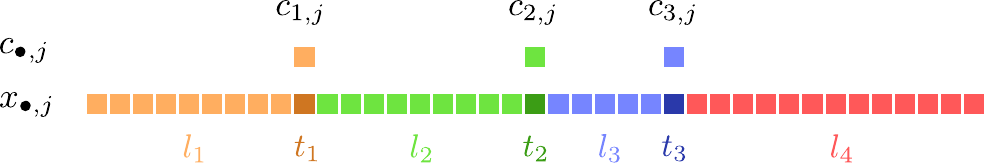}
\caption{Distribution of labels of $x_{i,j}$, determined by the position of boundaries $c_{k,j}$, for $N_b = 3$. Each label depends at most on two boundaries.}
\label{fig:notation_labels}
\end{figure}

The term $p(c|y)$ does not depend on $b$, so $q_b$ integrates out. Moreover, both $p(c|y)$ and $q_c$ factorize over columns. Hence we can rewrite the first summand of \eqref{eq:J-functional} as 
\begin{equation*}
\label{eq:first-summand-not-vectorized}
-\int_{b} \sum_{c} q_{b}(b) q_{c}(c) \log p(c|y) = -\sum_{j=1}^M \sum_{\col{c}{j}}\Big( q_c(\col{c}{j}) \sum_{i=1}^N \log p(x_{i,j}(\col{c}{j})|y_{i,j}) \Big)\,,
\end{equation*}
where the second sum ranges over all combinations of boundary assignments for $c_{\bullet,j}$. We can further simplify this equation by noting that each label $x_{i,j}$ depends at most on two $c_{k,j}$, as illustrated in Fig. \ref{fig:notation_labels}. This allows us to split the inner sum into $k+1$~sums, each summing over pixels with labels $l_k$ and $t_k$ or $l_{N_b+1}$ respectively, and sum out all $c_{k,j}$ independent of these labels. 

For each pair $(l_k,t_k)$ of labels we define matrices $\Psi_{k,j}$, whose entries equal the sum over pixel $y_{i,j}$ with $x_{i,j} \in \{t_k,l_k\}$
\begin{equation*}
(\Psi_{k,j})_{m,n} = \!\!\sum_{i=m+1}^{n-1} \beta^l  \log p(x_{i,j} = l_k|y_{i,j}) + \beta^t \log p(x_{n,j} = t_k|y_{n,j}) 
\end{equation*}
for $k=2,\ldots,N_b$, $j=1, \ldots, M$ and $1 \leq m < n \leq N$. Entries for $n \leq m$ are not defined and set to negative infinity. Accordingly, we introduce vectors $(\psi_{1,j})_n$ and $(\psi_{N_b,j})_n$, representing sums over pixels with labels $l_1,t_1$ and $l_{N_b+1}$ depending on $c_{1,j}$ and $c_{N_b,j}$ respectively. We now can write
\begin{equation*}
\sum_{c_{k-1,j}} \sum_{c_{k,j}} \qcjoint (\Psi_{k,j})_{c_{k-1,j},c_{k,j}} = \langle (\qcjointpure)^T,\Psi_{k,j}\rangle
\end{equation*}
where $\langle A,B \rangle$ denotes the trace of $AB$ and determines the form of $\qcjointpure$.
Finally, we write the first summand of \eqref{eq:J-functional} in vector form:
\begin{equation}
\label{eq:vectorized-first-term}
-\sum_{j=1}^M \Big((q_{c;1,j})^T \psi_{1,j} + \sum_{k=2}^{N_b} \langle (\qcjointpure)^T,\Psi_{k,j}\rangle + q_{c;N_b,j}^T \psi_{N_b,j} \Big) \,.
\end{equation}

\subsection{Second Summand $\log p(c|b)$ of $J(q_b,q_c)$}
\label{chap:p(c|b)}
The second term $p(c|b)$ depends on $q_c$ \textit{and} $q_b$, so we have to take care of both expected distributions. We start out this section by making explicit the expectation with respect to $q_b$. 
As a prerequisite, we define the two marginal distributions of $p(b)$ introduced in~\eqref{eq:c-given-b}. Using the simplifying notation $p(b_{j}|b_{\setminus j}) = p(\col{b}{j}|\col{b}{\setminus j})$, by the standard rule for conditional Gaussian distributions~\citep[p.~200]{rasmussen2006} we obtain
\begin{equation}
\begin{gathered}
\label{eq:conditional-sec-summand}
p(b_{j}|b_{\setminus j}) = \mc{N}(b_{j};\mu_{j|\setminus j},\Sigma_{j|\setminus j})\\
\mu_{j|\setminus j} = \mu_{j} - \Sigma_{j|\setminus j} K_{j,\setminus j} 
(b_{\setminus j} - \mu_{\setminus j}),\qquad
\Sigma_{j|\setminus j} = (K_{jj})^{-1} \,,
\end{gathered}
\end{equation}
the marginal distribution of the $N_b$ boundary positions in column $j$, conditioned on the $N_b(M-1)$ remaining boundary positions $b_{\setminus j}$. The 1-dimensional densities for $b_{k,j}|b_{\sm j}$ are obtained by marginalizing over~\eqref{eq:conditional-sec-summand}, with mean $(\mu_{j|\setminus j})_{k}$ and variance $(\Sigma_{j|\setminus j})_{k,k}$. 

Similarly, we define $p(b_{k,j}|b_{k-1,j})$
the density of boundary position~$b_{k,j}$ given the position of the neighboring boundary $k-1$ in column $j$. Its mean $\mucond$ and variance $\sigmacond$ are calculated in the same fashion as in \eqref{eq:conditional-sec-summand}.
We now can express the probabilities $p(c_{1,j}|b)$ and $p(c_{k,j}|c_{k-1,j},b)$ introduced in~\eqref{eq:c-given-b}
in terms of integrals 
\begin{equation*}
\begin{aligned}
\label{eq:integral-p-c-b}p(c_{k,j} = n|c_{k-1,j} = m,b) = \\
\int_{n-\frac{1}{2}}^{n+\frac{1}{2}} \int_{m-\frac{1}{2}}^{m+\frac{1}{2}}  \mc{N}(\tau;(\mu_{j|\setminus j})_{k},(\Sigma_{j|\setminus j})_{k,k}) 
& \mc{N}(\tau;\mucond,\sigmacond) \, d\tau d\nu.
\end{aligned}
\end{equation*}
and accordingly for $p(c_{1,j}=n|b)$\footnote{Note that the dependency on $\nu$ is contained in $\mucond$ as $b_{k-1,j}$.}.
These terms depend on $b_{\setminus j}$ through $(\mu_{j|\setminus j})_{k}$, hence on $q_b$ too.
It suffice to adopt the most crude numerical integration formula (integrand $=$ step function) in order to make this dependency explicit: 
$\int_{a-1/2}^{a+1/2} f(x) dx \approx f(a)$.

Applying the logarithm to $p(c|b)$, we obtain a representation that is convenient for $\int_{b} \dotsb q_{b} db$. 
We defined $q_b$ as a Gaussian distribution \eqref{eq:q-b-def}, therefore
the moments of $b_{\sm j}$ with respect to $q_b$ are given by
\begin{equation}
\label{eq:moments}
\EE_{q_b}[b_{\setminus j}] = \bar{\mu}_{\setminus j} \qquad \mathrm{and} \qquad \V_{q_b}[b_{\setminus j}] = \bar{\Sigma}_{\setminus j, \setminus j} + \bar{\mu}_{\setminus j} \bar{\mu}_{\setminus j}^T \,.
\end{equation}
As a result, we established all necessary prerequisites to write the terms $\mb{E}_{q_b}[\log p(c_{1,j}|b)]$ and $\mb{E}_{q_b}[\log p(c_{k,j}|c_{k-1,j},b)]$ in an explicit form, that is suitable for an optimization with respect to $\bar{\mu}$ and $\bar{\Sigma}$. Details are provided in \ref{app:expectation}.

We now address the expectation with respect to $q_c$. Similar arguments as for $p(c|b)$ hold for $p(c|y)$ too: We can split the sum over $c_{\bullet,j}$ into parts depending (at most) on two neighboring boundaries $c_{k-1,j}$ and $c_{k,j}$.  We define matrices $\Omega_{k,j}$ as
\begin{equation*}
\nonumber (\Omega_{k,j})_{m,n} = \,\mb{E}_{q_b}[\log p(c_{k,j} = n|c_{k-1,j} = m,b)],
\end{equation*}
for $k=2,\ldots,N_b$, $j=1, \ldots, M$ and $1 \leq m < n \leq N$, and vectors $(\omega_{1,j})_{n}$ for terms $\mb{E}_{q_b}[\log p(c_{1,j} = n|b)]$ accordingly. Finally, we can write the expectation of the second term in vectorized form as
\begin{equation}
\label{eq:vectorized-second-term}
-\sum_{j=1}^M \Big( (q_{c;1,j})^T \omega_{1,j} + \sum_{k=2}^{N_b} \langle (\qcjointpure)^T,\Omega_{k,j} \rangle\Big) \,.
\end{equation}

\begin{figure}[!t]
\centerline{
\subfloat[]{\includegraphics[width=0.32\textwidth]{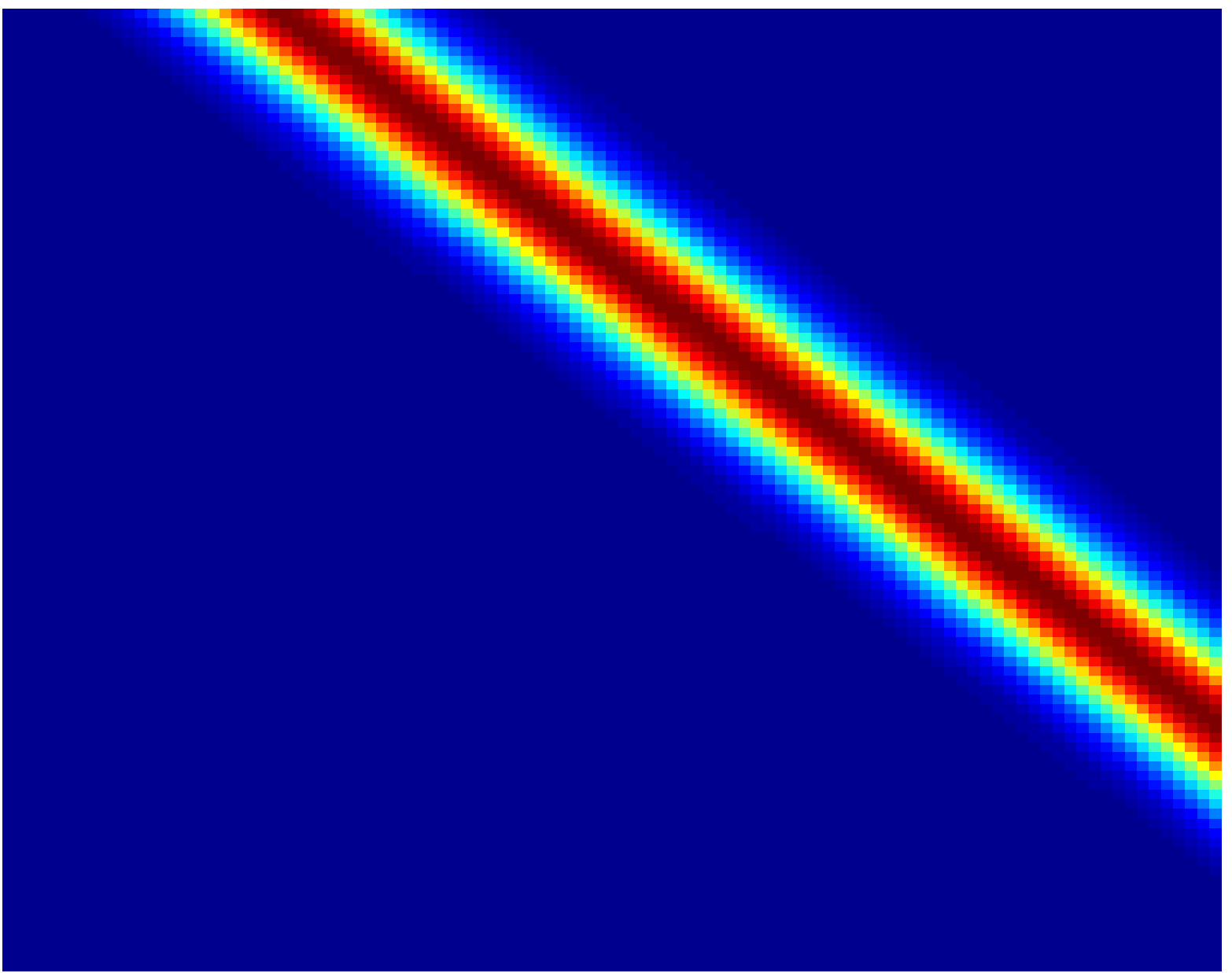}}
\hfill
\subfloat[]{\includegraphics[width=0.32\textwidth]{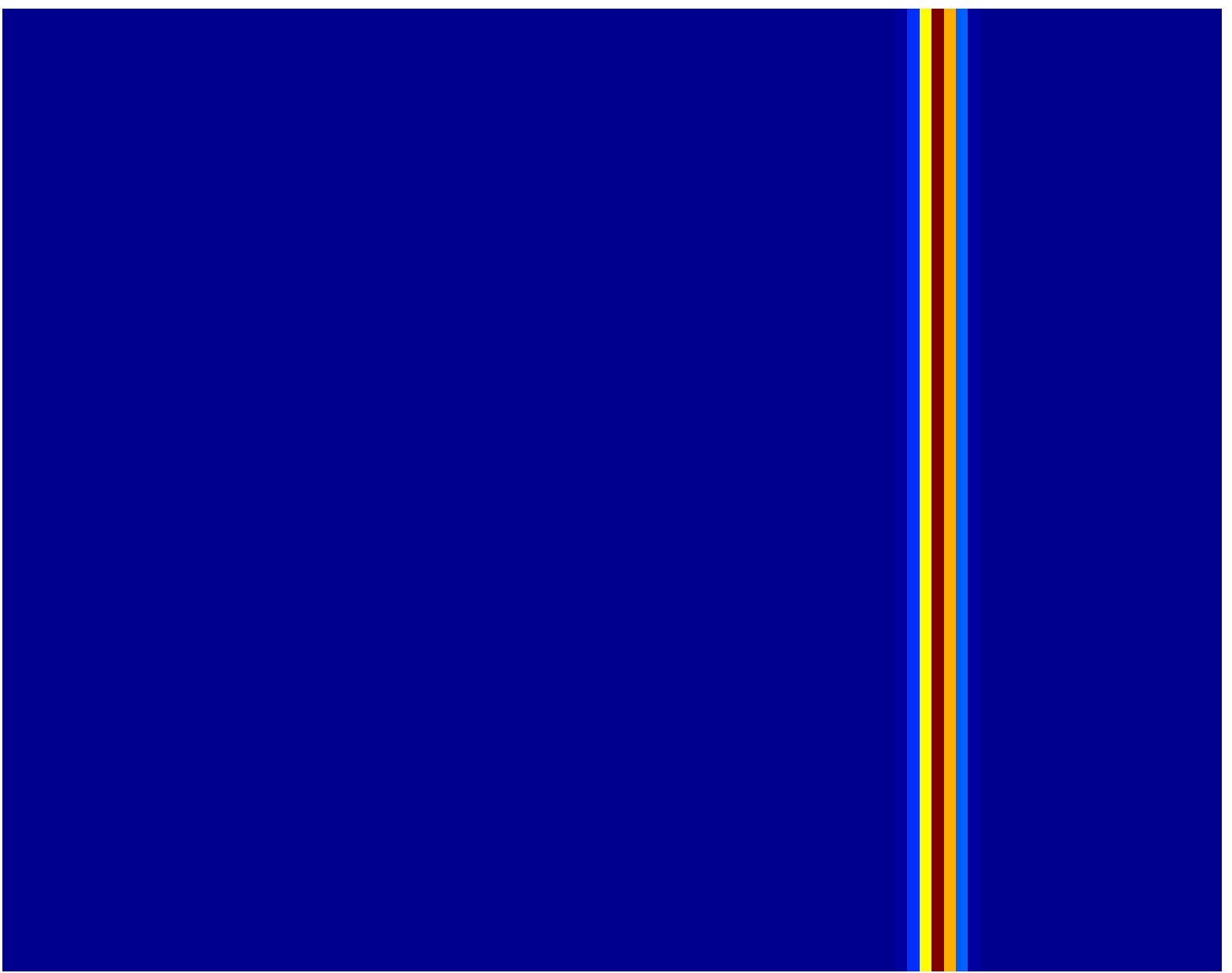}}
\hfill
\subfloat[]{\includegraphics[width=0.32\textwidth]{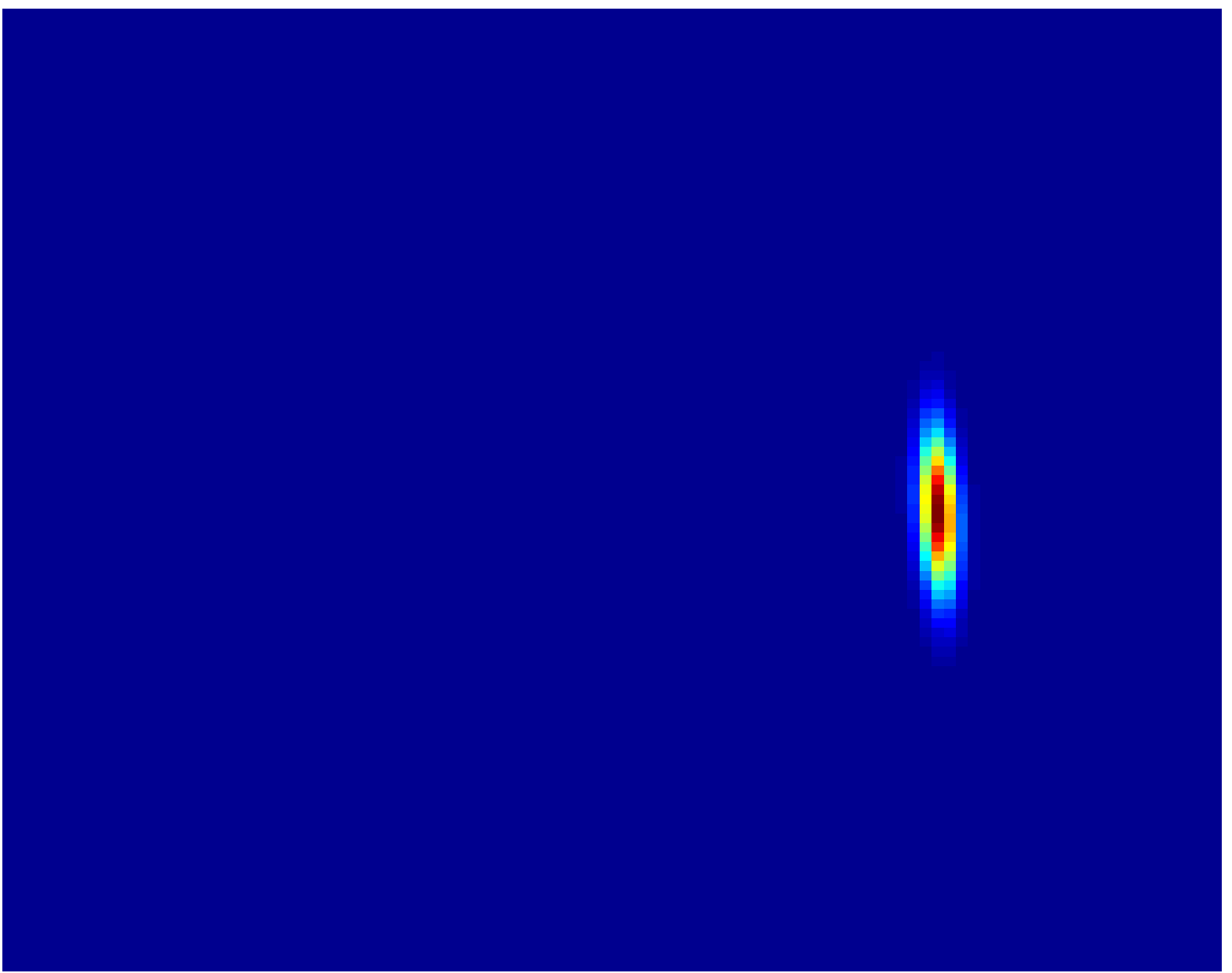}}
}
\caption{Illustration of a transition matrix $\exp (\Omega_{k,j})$ (c) and the local (a) and global (b) shape information it is composed of. The plots show the exponential version that is used during the optimization of $q_c$ (see Section \ref{chap:opt-q-c}), in order to illustrate the inherent sparsity that we utilize to speed up the calculation of~$q_c$.}
\label{fig:transitionDetail}
\end{figure}

Fig. \ref{fig:transitionDetail} shows a transition matrix $\Omega_{k,j}$
(c) and it's two components {$\mb{E}_{q_b}[\log p(b_{k,j}|b_{k-1,j})]$} (a) and {$\mb{E}_{q_b}\big[\log p(b_{k,j}|b_{\sm{j}})\big]$} (b) for $m,n = 101,\ldots,200$. Since the sum-product algorithm used to find the optimal $q_c$ (see Section \ref{chap:opt-q-c}) requires $\exp(\Omega_{k,j})$, our plots show the exponential version too, in order to illustrate the inherent sparsity. We see how $\Omega_{k,j}$ is build by combining prior information about the relative distance between $b_{k,j}$ and $b_{k-1,j}$~(a) with the distribution of $b_{k,j}$ conditioned on information from all other columns via $\mb{E}_{q_b}[b_{\setminus j}] = \bar{\mu}_{\setminus j}$~(b).

\subsection{Third Summand $\log p(b)$ of $J(q_b,q_c)$}
Concerning the third summand, $q_c$ sums out. Rewriting the Gaussian using the trace and making the expectation explicit, we obtain
\begin{equation}
- \int_b q_b(b) \log p(b) db = C + \frac{1}{2}\langle K, \bar{\Sigma} + \bar{\mu}\bar{\mu}^T - 2\bar{\mu}\mu^T + \mu\mu^T \rangle
\end{equation}
i.e.~a function depending on the parameters $\bar{\mu}$ and $\bar{\Sigma}$ of $q_{b}$.


\subsection{Entropy Terms of $J(q_b,q_c)$}
Finally, we make explicit the negative entropy of $q_b$ and $q_c$.
\begin{align}
-H_{q_b}(b) &= \int_b q_b(b) \log q_b(b) db = C -\frac{1}{2}\log|\bar{\Sigma}| \,. \\
\label{eq:entropy-q-c}-H_{q_c}(c) &= \sum_{j=1}^M \bigg(\sum_{k=1}^{N_b}\sum_{c_{k,j}} q_{c}(c_{k,j}) \log q_{c}(c_{k,j})  \\
\nonumber &\qquad + \sum_{k=2}^{N_b}\sum_{c_{k-1,j},c_{k,j}} \qcjoint \log\frac{\qcjoint}{q_{c}(c_{k-1,j})q_{c}(c_{k,j})}\bigg)
\end{align}
The first summand of $H_{q_c}$ is comprises singleton entropies whereas the second one comprises the mutual information \citep[p.~19]{cover2006} between the random variables $c_{k,j}$ and $c_{k-1,j}$.

\subsection{Explicit Formulation of the Objective Function $J(q_b,q_c)$}
Combining all terms, we can reformulate \eqref{eq:J-functional} into a functional that can be optimized with respect to $q_c$ and the parameters $\bar{\mu}$ and $\bar{\Sigma}$ of $q_b$
\begin{equation}
\label{eq:complete_functional}
\begin{aligned}
\min_{q_c,\bar{\mu},\bar{\Sigma}} & -\Big(\sum_{j=1}^M (q_{c;1,j})^T\theta_{1,j} + \sum_{k=2}^{N_b} \langle(\qcjointpure)^T \Theta_{k,j}\rangle + (q_{c;N_b,j})^T\theta_{N_b,j}\Big) \\
& \qquad + \frac{1}{2}\langle K, \bar{\Sigma} + \bar{\mu}\bar{\mu}^T - 2\bar{\mu}\mu^T \rangle - \frac{1}{2}\log|\bar{\Sigma}| - H_{q_c}(c) + C\\
\mathrm{s.\;t.} & \quad  q_{c;k,j} \in \Delta_N \quad k=1,\ldots,N_b, j = 1,\ldots,M  \qquad \mathrm{and} \qquad \eqref{eq:q-c-pairwise-constraints}
\end{aligned}
\end{equation}
where we combined the terms of \eqref{eq:vectorized-first-term} and \eqref{eq:vectorized-second-term} into $\theta_{k,j}$ and $\Theta_{k,j}$. Note that the necessary constraint 
$\bar{\Sigma} \in \mc{S}_{++}$ 
is implicitly given by the logarithmic barrier term $\log|\bar{\Sigma}| = \sum_i \log\bar{\lambda}_i$, where $\bar{\lambda}_i$ is the $i$-th eigenvalue of $\bar{\Sigma}$,
hence it has not to be enforced explicitly.

%% file: mod_var_opt.tex
We alternatingly optimize the objective function \eqref{eq:complete_functional} with respect to $q_c$ and the parameters of $q_b$.
Optimization of $q_{c}$ corresponds to inference of chain graphs and can be accomplished by the sum-product algorithm \citep[p.~402]{bishop2006}, whereas the optimization of $q_{b}$ can be done in closed form.
Both subproblems are convex, thus by alternatingly optimizing with respect to $q_b$ and $q_c$, the functional $J(q_b,q_c)$, being bounded from below over the feasible set of variables, is guaranteed to converge to some minimum. 

\subsection{Optimization of $q_c$}
\label{chap:opt-q-c}

Fixing all terms in \eqref{eq:complete_functional} that depend on parameters of $q_b$, we obtain an optimization problem that can be split into column-wise \emph{convex} subproblems, since each is composed solely of linear terms and the negative entropy of a chain graph, subject to simplex constraints.
Making the constraints explicit using Lagrange multipliers, and derivating 
with respect to all $q_{c;k,j}$, we obtain a set of update equations which can be shown to correspond to sum-product updates \citep[p.~83]{wainwrigth2008}. Thus iteratively optimizing $q_c$ for each column $j$ is guaranteed to converge to some fix point $q_c^{\mathrm{opt}}$, which corresponds to the global optimum.


\subsection{Optimization of $q_b$}
\label{chap:optimization-q-b}
Considering in \eqref{eq:complete_functional} only terms depending on $\bar{\Sigma}$, we obtain the optimization problem
\begin{equation}
\label{eq:optimize-q-b-Sigma}
\min_{\bar{\Sigma}} -\frac{1}{2} \log|\bar{\Sigma}| + \frac{1}{2} \la K + \tilde{P},\bar{\Sigma} \ra 
\end{equation}
which has the closed-form solution: $\bar{\Sigma}_{\mathrm{opt}} = (K + \tilde{P})^{-1}$. The newly introduced matrix~$\tilde{P}$ contains the dependencies on $\bar{\Sigma}$ of terms $\omega_{1,j}$ and $\Omega_{k,j}$. Being independent of $q_c$, we only have to calculate it once. Furthermore, since it is composed out of linear combinations of submatrices of $K$, it can be expressed implicitly in terms of $W$ and $\sigma^2I$. Details are provided in \ref{app:optimization}. 

For $\bar{\mu}$, we obtain
\begin{equation}
\label{eq:optimize-q-b-mu}
\min_{\bar{\mu}} \frac{1}{2} \la K + \tilde{P},\bar{\mu}(\bar{\mu} - 2\mu)^T \ra + \tilde{p}^T\bar{\mu}
\end{equation}
and $\bar{\mu}_{\mathrm{opt}} = \mu - \tilde{p}^T\bar{\Sigma}_{\mathrm{opt}}$. Again details for $\tilde{p}$, concerning the dependencies of $\omega_{1,j}$ and $\Omega_{k,j}$, are given in \ref{app:optimization}. To minimize \eqref{eq:optimize-q-b-mu}, we use conjugate gradient descent which enables us to calculate $\bar{\mu}_{\mathrm{opt}}$ using $(K + \tilde{P})$ instead of $(K + \tilde{P})^{-1}$.

\subsection{Initialization}
We start the optimization of \eqref{eq:complete_functional} by initializing the distribution $q_c$. This is done by setting distributions $p(b_{k,j}|b_{\setminus j})$ to a uniform distribution, since we yet lack the distribution $q_b$. Afterwards, we can initialize $q_b$ via \eqref{eq:optimize-q-b-Sigma} and \eqref{eq:optimize-q-b-mu}. Subsequently, we iterate both optimizations alternatingly until $J(q_b,q_c)$ converges.

%% file: res_setup.tex
\subsection{Data Acquisition}
\label{sec:data-sets}
\begin{figure}
\centerline{
\subfloat[2-D circular scan]{\includegraphics[width=2in]{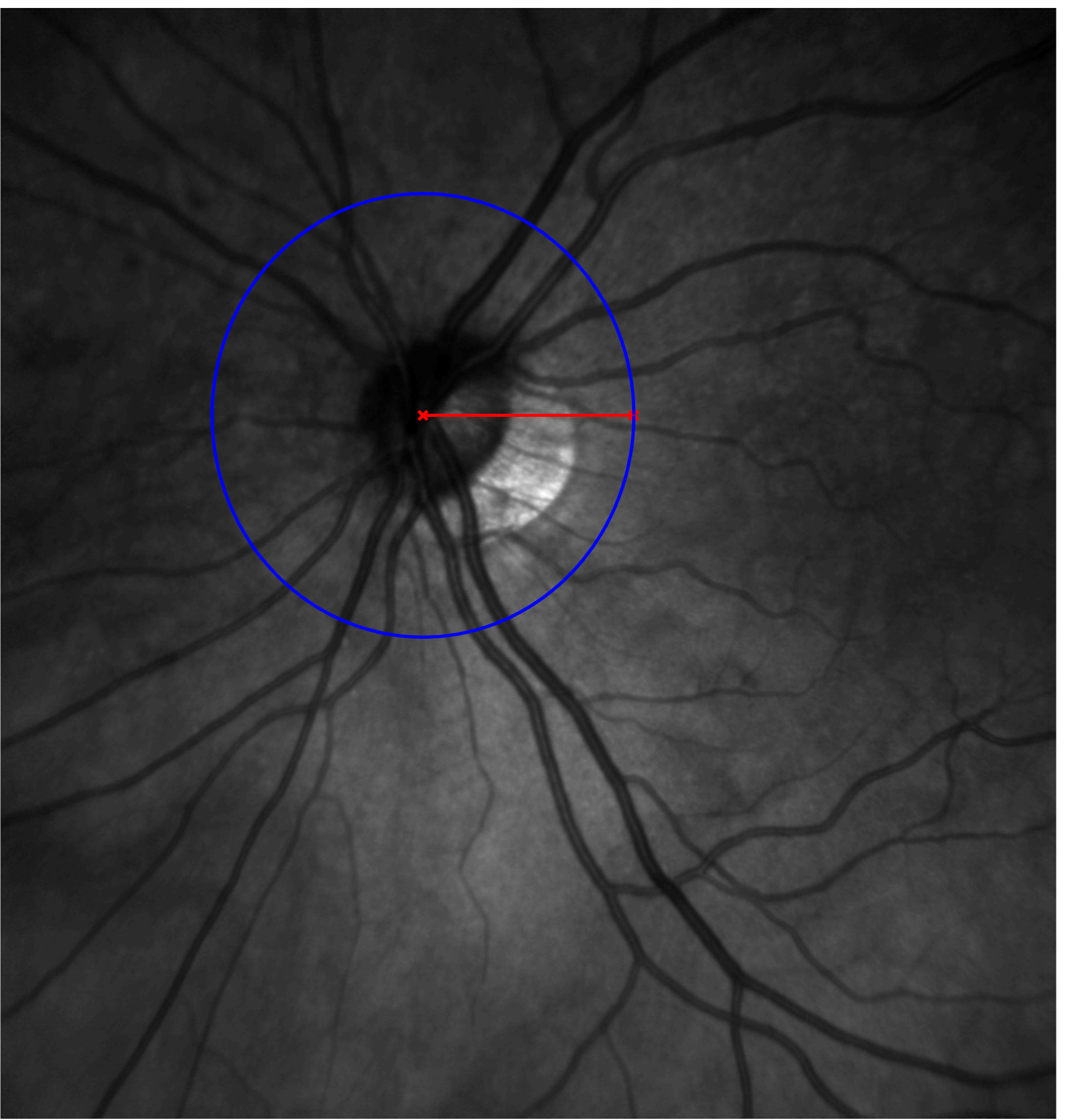}}
\hspace{10mm}
\subfloat[3-D volume]{\includegraphics[width=2in]{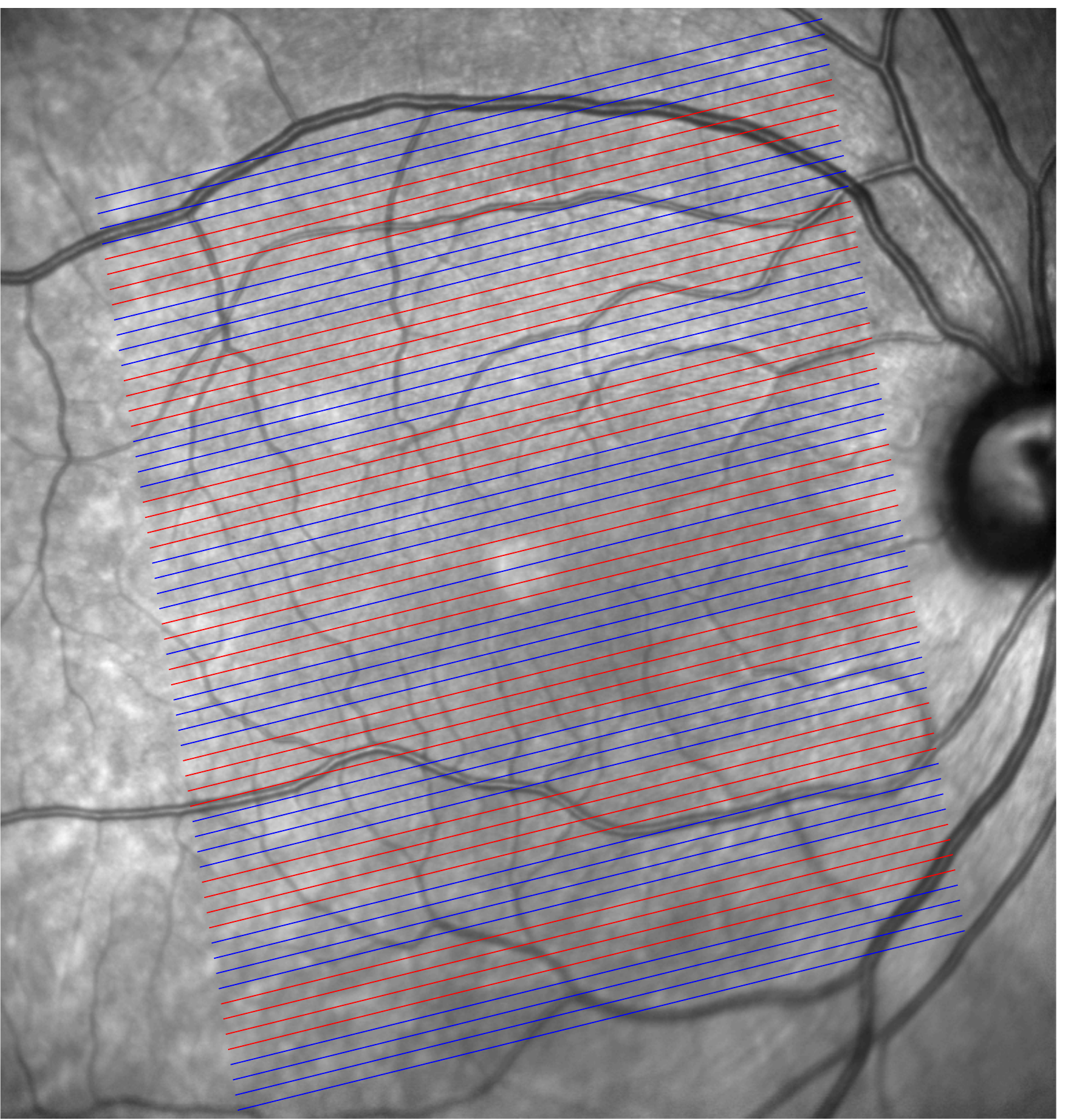}}
}
\caption{SLO fundus images that exemplarily depict (a) the trajectory and radius of a 2-D circular scan centered around the optical nerve head and (b) the area covered by a 3-D volume consisting of 61 B-Scans centered at the fovea. Different colors illustrate the partitioning into 17 different regions. }
\label{fig:SLO}
\end{figure}

Circular B-scans measured around the optic nerve head were acquired from 80 healthy as well as from 66 glaucomatous subjects using a Spectralis HRA+OCT device (Heidelberg Engineering, Germany). Each scan had a diameter of 12$^{\circ}$, corresponding to approximately $3.4\,mm$, and consisted of $M=768$ A-scans of depth resolution $3.87\mu{}m$/pixel ($N=496$ pixels), see Fig.~\ref{fig:SLO}~(a). Ground truth for the crucial boundary separating NFL and GCL as well as a grading for the pathological scans was provided by a medical expert: \textit{pre-perimetric} glaucoma (PPG), meaning the eye is exhibiting structural symptoms of the disease but the visual field and sight are not impaired yet, as well as \textit{early}, \textit{moderate} and \textit{advanced} primary open-angle glaucoma (PGE, PGM and PGA). Ground truth for the remaining eight boundaries was produced by the first author. To measure interobserver variability, a second set of labels for the healthy B-scans was obtained by the second author.

The second data set consisted of fovea-centered 3-D volumes, acquired from 35 healthy subjects using the same device as above. Each volume was composed of $61$~B-Scans of dimension $500 \times 496$, covering an area of approximately $5.7 \times 7.3\,mm$. Ground truth was obtained as follows: Each volume was divided in 17 regions, and a B-scan randomly drawn from each region was labeled with the previously introduced nine boundaries. Fig.~\ref{fig:SLO}~(b) depicts the location of all 61 B-Scans and their partition into regions indicated by color. 

\subsection{Generative vs. Discriminative, Transition vs. Boundary Appearance Terms}
\label{sec:generative-vs-discriminative}
Following \eqref{eq:kl_minimization}, we described the introduction of discriminative appearance models as an alternative to generative ones. Furthermore, we introduced switches $\beta^l$ and $\beta^t$ in \eqref{eq:appearance-terms} to enable or disable layer and boundary appearance terms, respectively. This section explains why we settled for discriminative boundary terms.

\begin{figure}
\centerline{
\subfloat[$\beta^l,\beta^t = 1$, Gen.]{\includegraphics[width=1.6in]{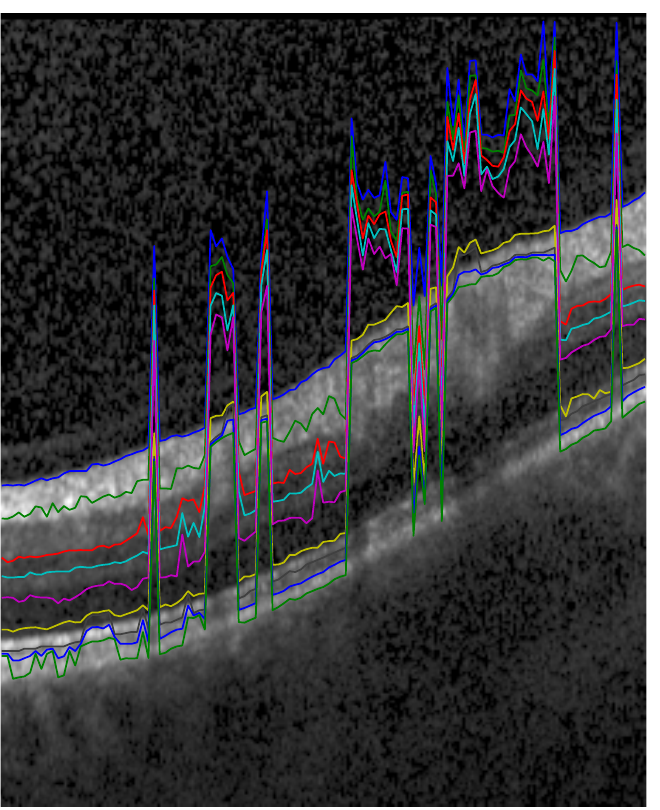}}
\hfill
\subfloat[$\beta^l = 0$, Gen.]{\includegraphics[width=1.6in]{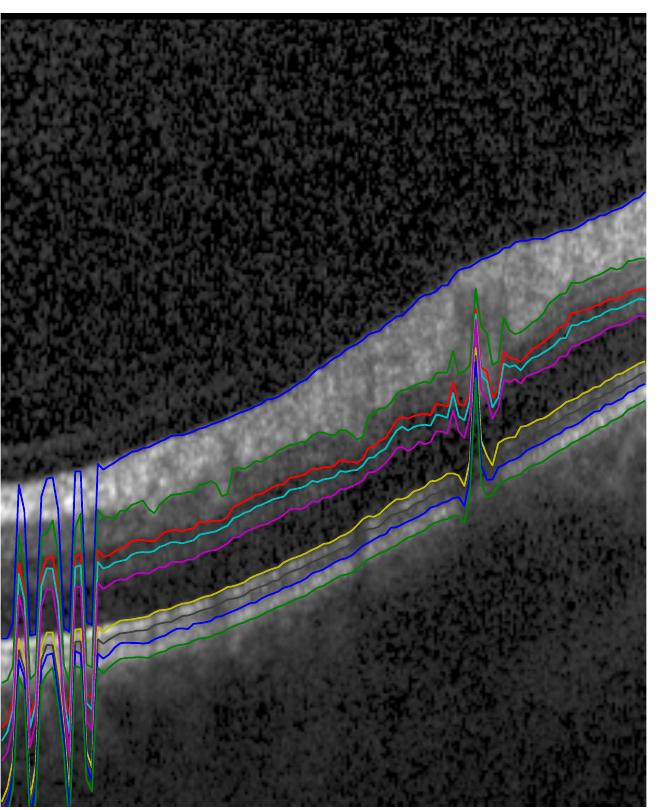}}
\hfill
\subfloat[$\beta^l = 0$, Disc.]{\includegraphics[width=1.6in]{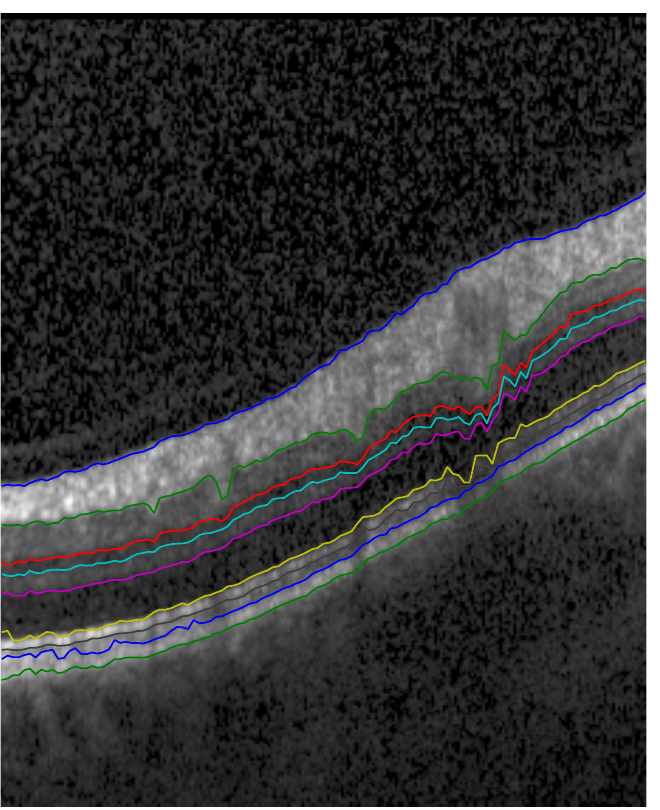}}
}
\caption{(a)-(c) Close-up view of initialization results for different configurations of appearance terms. Switches $\beta^l$ and $\beta^t$ include or exclude layer and transition appearance terms.}
\label{fig:configExamples}
\end{figure}

Using the set of healthy circular scans, we tested the model with generative layer as well as boundary terms, i.e. $\beta^t = \beta^l = 1$. This configuration turned out to be sensitive to distortions of the texture caused for example by blood vessels. The result were initializations above the actual retina, since the model misinterpreted the shaded area as parts of the choroid, as shown in Fig. \ref{fig:configExamples} (a). We then disabled the layer appearance terms, i.e.\ set $\beta^l = 0$. This solved the previous issue, but spuriously led to some columns being initialized below the retina, due to very high probabilities for some boundary classes caused by relatively small class model variances, i.e. narrow and steep normal distributions. For patches close to the mean, the probabilities for those classes happened to be up to $100$ times larger than for other classes. This caused false positive class responses in the choroid to displace the whole initialization for these columns, as displayed in Fig. \ref{fig:configExamples} (b).

Switching to discriminative probabilities solved this issue as well, since the local normalization limits all probabilities to $1$ and gives each appearance class the same influence. Thus false-positives did not possess the probability mass any more to displace the whole column segmentation, see Fig. \ref{fig:configExamples}~(c).
Notice that the layer terms, although switched off by setting $\beta^l = 0$, are utilized indirectly, since they contribute to the normalization of terms $p(x_{i,j}(c)|y_{i,j})$, see~\eqref{eq:discriminative-app-terms}. Thus strong layer appearance terms can rule out certain parts of the OCT scan for segmentation.

\subsection{Model Parameters}
\label{chap:parameters}

\begin{table}
\centering
\caption{Set of model parameter values used throughout all experiments.}
\label{tab:parameter-settings}
\begin{tabular}{c c c c c c c}
\toprule
 & \multicolumn{3}{c}{Appearance} & Shape & Inference \vspace{2.5pt}\\
Parameters & $\alpha_{\mathrm{glasso}}$ &  $q_{\mathrm{pca}}$ & Patch-Size & $q_{\mathrm{ppca}}$ & Variance of ~$p(b_{k,j}|b_{\sm j})$\\
\midrule
Value & $0.01$& $20$ &  $15\times 15$ & $20$ & $10$ \\
\bottomrule
\end{tabular}
\end{table}
Table \ref{tab:parameter-settings} summarizes the model parameters and how they were set. For the appearance model we set $\alpha_{\mathrm{glasso}}$ to $0.01$, which resulted in sparse covariance matrices  $\Sigma_{x_{i,j}}$ that speed up computations significantly. A patch-size of $15 \times 15$ and the projection onto the first $q_{\mathrm{pca}} =20$ eigenvectors resulted in smooth segmentation boundaries. Similar, we used $q_{\mathrm{ppca}} = 20$ eigenvectors to build the shape prior model, after examining the eigenvalue spectrum.

An important parameter during the inference is the variance of~$p(b_{k,j}|b_{\sm j})$: it balances the influence of appearance and shape. Artificially increasing this parameter results in broader normal distributions (cf. Fig. \ref{fig:transitionDetail} (b)), that allows~$q_c$ to take into account more observations around the mean of~$p(b_{k,j}|b_{\sm j})$. At the same time the influence of the appearance terms on $q_b$ is reduced, which results in a smoother mean~$\bar{\mu}$. Thus increasing the variance loosens the coupling between~$q_c$ and~$q_b$ and vice versa. A ten-fold increased variance turned out to provide a good balance between local appearance terms and shape regularization as well as between run-time and prediction accuracy.

We used the very same set of parameter values for all our experiments and performed no fine tuning separately for each data set. Hence it is plausible to assume that these values perform well on a broad range of data sets.

\subsection{Error Measures and Test Framework}
\label{chap:error_measures}
For each boundary we computed the unsigned distance $E^k_{\mathrm{unsgn}}$ in $\mu m$ between estimates $\hat{c}_{k,j} = \mb{E}_{q_c}[c_{k,j}]$ and manual segmentations $\tilde{c}_{k,j}$ (ground truth) as
\begin{equation*}
E^k_{\mathrm{unsgn}} = M^{-1}\sum_{j=1}^{M} |\hat{c}_{k,j} - \tilde{c}_{k,j}|,\qquad E_\mathrm{unsgn} = N_b^{-1} \sum_{k=1}^{N_b} E^k_{\mathrm{unsgn}}\,.
\end{equation*}
For volumes we additionally have to average over regions. For each data set, we provide the mean error $\bar{E}_\mathrm{unsgn}$ and it's standard deviation (SD) $\sigma_{E_\mathrm{unsgn}}$. 

Results were obtained via cross-validation: After splitting each data set into a number of subsets, each subset in turn is used as a test set, while the remaining subsets are used for training. This provides an estimate of the ability to segment new (unseen) test scans.
We used 10-fold cross-validation for the set of non-pathological circular scans and leave-one-out cross-validation for the volumes, to maximize the number of training examples in each split. For the set of glaucomatous scans, we used a single model trained on all healthy scans.

%% file: res_imp.tex
We implemented our approach in MATLAB. \rmv{For prediction we first require data terms $p(c|y)$ for each pixel and all classes of appearance models. Next we infer an initial solution for $q_c$.  With this we initialize $q_b$ and then iterate the optimization of $q_c$ and $q_b$ until convergence of the segmentation.}
The main bottle-neck, the sum-product algorithm used to find an optimal solution for $q_c(c)$, was implemented in C and incorporated into MATLAB via the Mex-interface. To further decrease running time, we exploited the inherent sparsity of the transition matrices $\Omega_{k,j}$, as illustrated in Fig.~\ref{fig:SLO}. Also, wherever possible we transferred expensive matrix-vector multiplications to the GPU, using a wrapper for MATLAB called GPUmat \cite{messmer2008}. 
Segmenting all 61 B-Scans of a 3-D volume took~\unit[60]{s}, with memory requirements of about \unit[2]{GB}, measured on a Core i7-2600K 3.40GHz. 

%% file: res_circ.tex
\renewcommand{\arraystretch}{1.1}
{\setlength{\tabcolsep}{0.12cm}
\begin{table}
\centering
\caption{Results in $\mu m\!\pm\!\mathrm{SD}$ ($3.87\mu m\,\widehat{=}\,1px$) for 2-D circular scans (separately for healthy eyes as well as the different degrees of glaucoma, pre-perimetric, early, moderate and advanced) and 3-D scans of healthy subjects. Numbers within brackets denote the respective data set size.}
\label{tab:performance-circ-scans}
\begin{tabular}{c c >{\hspace{0.4pc}}c c c c>{\hspace{0.4pc}} c}
\toprule
& 2-D Healthy & \multicolumn{4}{c}{2-D Glaucoma} & 3-D Healthy\\
$k$ & \textit{All (80)} & \textit{PPG (22)}& \textit{PGE (22)}& \textit{PGM (13)}& \textit{PGA (9)} & \textit{All (35)}\\
\midrule
1 & $2.06\!\pm\!0.57$& $2.60\!\pm\!0.85$& $3.76\!\pm\!1.42$& $4.51\!\pm\!1.18$& $6.53\!\pm\!2.76$& $1.36\!\pm\!0.18$\\
2 & $4.68\!\pm\!1.13$& $6.66\!\pm\!2.41$& $5.65\!\pm\!1.66$& $6.74\!\pm\!1.64$& $9.95\!\pm\!4.74$& $3.32\!\pm\!0.37$\\
3 & $3.67\!\pm\!0.84$& $4.57\!\pm\!1.18$& $5.37\!\pm\!1.33$& $5.49\!\pm\!1.00$& $8.80\!\pm\!3.03$& $3.17\!\pm\!0.44$\\
4 & $3.31\!\pm\!0.78$& $4.43\!\pm\!1.09$& $5.78\!\pm\!1.48$& $5.44\!\pm\!1.19$& $8.30\!\pm\!2.21$& $3.23\!\pm\!0.56$\\
5 & $3.30\!\pm\!0.75$& $4.34\!\pm\!1.63$& $4.40\!\pm\!1.14$& $4.15\!\pm\!0.68$& $5.05\!\pm\!0.92$& $3.27\!\pm\!0.66$\\
6 & $2.10\!\pm\!0.76$& $2.67\!\pm\!1.37$& $2.76\!\pm\!0.97$& $2.88\!\pm\!1.62$& $2.99\!\pm\!1.92$& $1.61\!\pm\!0.23$\\
7 & $2.34\!\pm\!1.05$& $2.59\!\pm\!1.11$& $2.95\!\pm\!1.27$& $2.21\!\pm\!0.68$& $2.42\!\pm\!0.44$& $1.86\!\pm\!0.32$\\
8 & $2.81\!\pm\!1.42$& $2.82\!\pm\!1.00$& $3.40\!\pm\!1.22$& $2.94\!\pm\!1.40$& $4.19\!\pm\!1.97$& $2.27\!\pm\!0.40$\\
9 & $2.01\!\pm\!1.14$& $2.06\!\pm\!0.65$& $1.63\!\pm\!0.48$& $1.64\!\pm\!0.25$& $2.36\!\pm\!1.18$& $2.07\!\pm\!0.48$\\
\midrule
$\bm{\varnothing}$ & $\textbf{2.92}\!\bm{\pm}\!\textbf{0.53}$& $\textbf{3.64}\!\bm{\pm}\!\textbf{0.68}$& $\textbf{3.97}\!\bm{\pm}\!\textbf{0.73}$& $\textbf{4.00}\!\bm{\pm}\!\textbf{0.53}$& $\textbf{5.62}\!\bm{\pm}\!\textbf{1.25}$& $\textbf{2.46}\!\bm{\pm}\!\textbf{0.22}$\\
\bottomrule
\end{tabular}
\end{table}

Average boundary-wise results are summarized in Table \ref{tab:performance-circ-scans}. 
In general, boundaries 1 and 6 to 9 turned out to be easier to segment than boundaries~2~to~5. For boundary 1 this stems from easily detectable textures, whereas boundaries 6-9 with their regular shape profit disproportionately from regularization by the shape prior. Boundaries 2-5 on the other hand pose a harder challenge with their high variability of texture and shape. The upper row in Fig.~\ref{fig:circ-examples} shows an example close to the average segmentation performance with~$E_\mathrm{unsgn} = 2.97\,\mu m$. 

\begin{figure}
\centerline{
\subfloat{\includegraphics[width=0.49\textwidth]{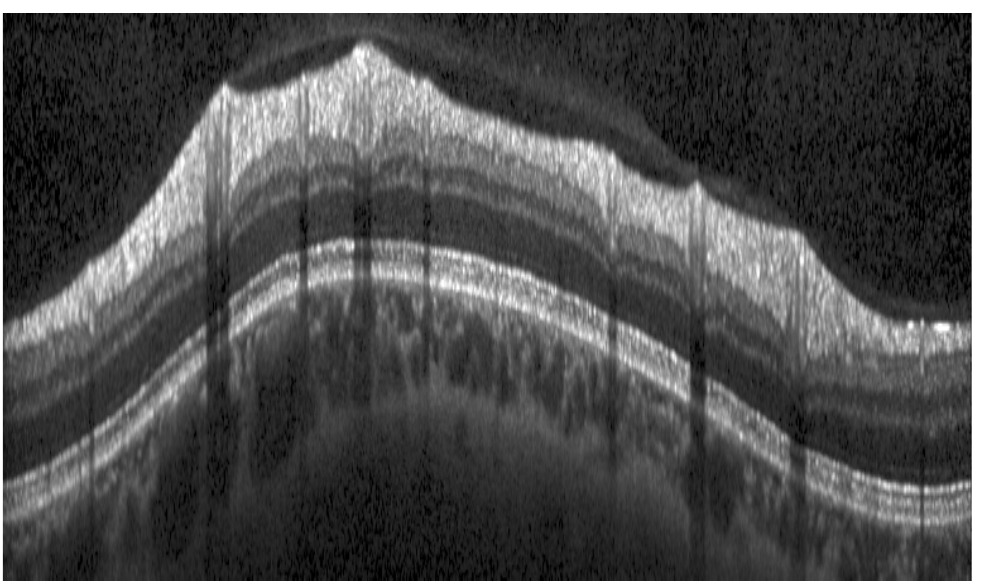}}
\hfill
\subfloat{\includegraphics[width=0.49\textwidth]{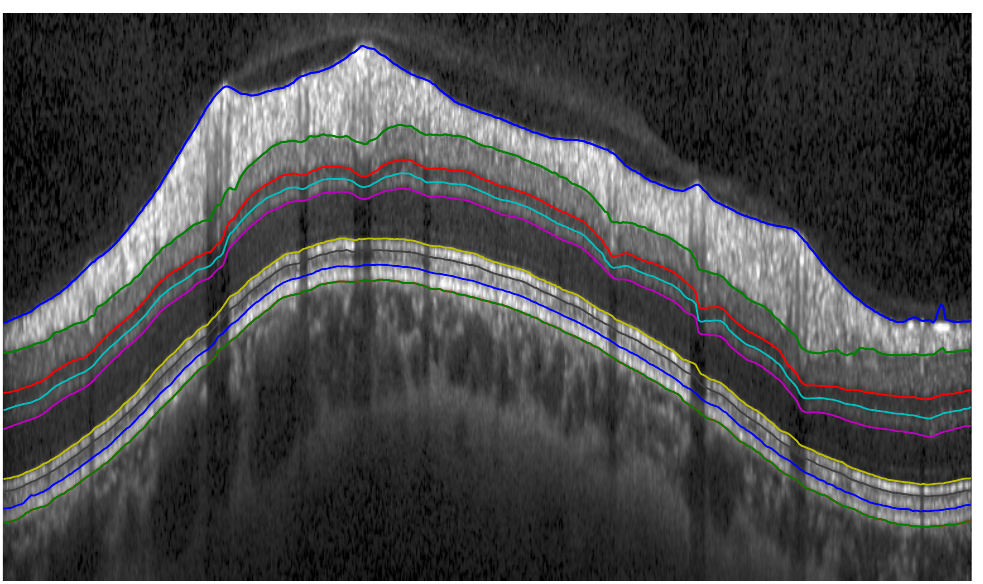}}
}
\vspace{-0.2cm}
\centerline{
\subfloat{\includegraphics[width=0.49\textwidth]{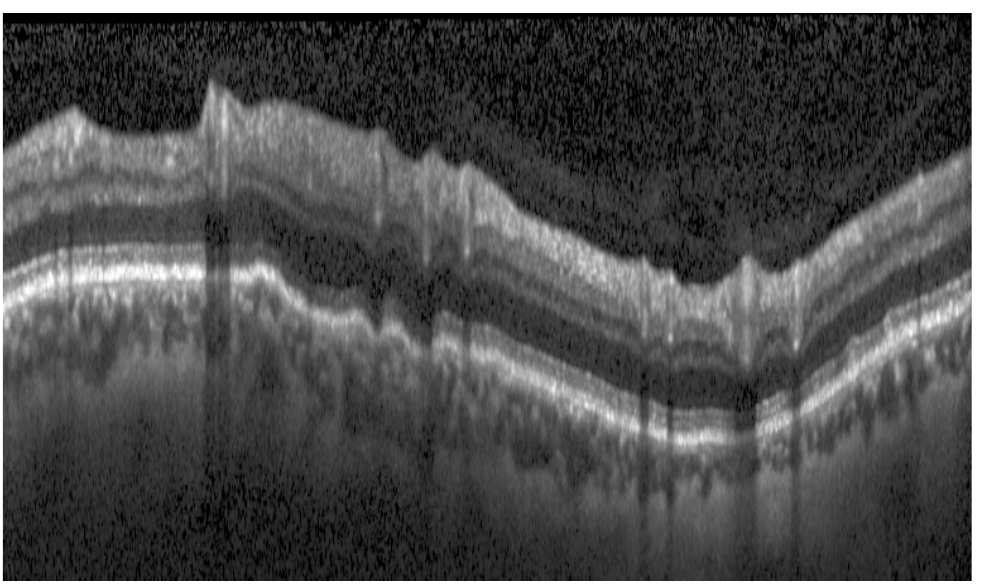}}
\hfill
\subfloat{\includegraphics[width=0.49\textwidth]{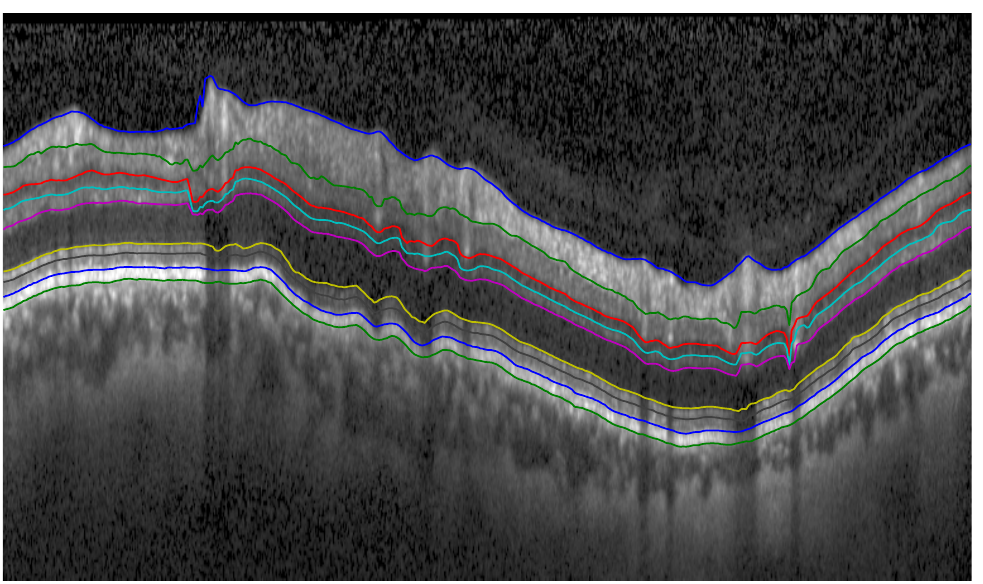}}
}
\caption{Top: Segmentation ($E_\mathrm{unsgn} = 2.97\,\mu m$) of a non-pathological circular scan. Bottom: Segmentation ($E_\mathrm{unsgn} = 5.09\,\mu m$) of an  advanced glaucomatous scan.}
\label{fig:circ-examples}
\end{figure}

For the pathological scans segmentation performance was comparable to the healthy scans, but decreased with the progression of the disease. This happened for two reasons: Since glaucoma is known to cause a thinning of the nerve fiber layer (NFL) \cite{schuman1995,bowd2001}, the shape prior trained on healthy scans may encounter difficulties adapting to very abnormal glaucomatous shapes. Furthermore, we observed a reduced scan quality for glaucomatous scans, also reported by others \cite{ishikawa2005,stein2006,mayer2010}, which in turn reduced the quality of the data terms\removed{~$p(y|c)$}. 
For advanced primary open-angle glaucoma, the NFL can even vanish at some locations. The appearance model for this layer, trained on healthy data, is not able to detect these extreme anomalies, which resulted in a comparatively low performance for some scans. We discuss possible modifications to overcome this problem in Section \ref{chap:discussion}. 

The bottom panels in Fig.~\ref{fig:circ-examples} show an example of a PGA-type scan and its segmentation. The scan exhibits the discussed reduced scan quality. Furthermore, the segmentation proves that the shape model can generalize well to pathological shapes as well as scan artifacts.
\subsubsection{Interobserver Variability}
A second set of labels was created for the healthy circular B-scan data set by the second author. For training and testing we utilized the same set-up as described earlier (10-fold cross-validation, parameters as in Table \ref{tab:parameter-settings}), but used the average of both labelings for training. In Table \ref{tab:interobserver-variability} we compare the predicted segmentations with the two labelings individually and with their average. Furthermore, we report the average absolute distance between both observers, the interobserver variability.

We see, that the resulting prediction errors are well within the range of the interobserver variability. The performance using the averaged labels improves compared to the case when using only one set of labels, c.f. first column of Table \ref{tab:performance-circ-scans}. This suggests an increased robustness of the averaged ground truth towards scan artifacts, ambiguous image regions and labeling bias.

\renewcommand{\arraystretch}{1.1}
{\setlength{\tabcolsep}{0.12cm}
\begin{table}
\centering
\caption{Interobserver variability as well as prediction performance of our segmentation approach compared to ground truth of observer 1 and 2 for the set of 80 healthy circular scans ($\mu m\!\pm\!\mathrm{SD}$ ($3.87\mu m\,\widehat{=}\,1px$)). The algorithm was trained on the averaged ground truth.}
\label{tab:interobserver-variability}
\begin{tabular}{c c c c c}
\toprule
& \begin{tabular}[x]{@{}c@{}}Obs.1 vs.\\Obs.2\end{tabular}& \begin{tabular}[x]{@{}c@{}}Algo. vs.\\Obs.1\end{tabular}& \begin{tabular}[x]{@{}c@{}}Algo. vs.\\Obs.2\end{tabular}& \begin{tabular}[x]{@{}c@{}}Algo. vs.\\Avg. Obs.\end{tabular}\\
\midrule
1 & $2.86\!\pm\!0.46$& $2.35\!\pm\!0.62$& $3.69\!\pm\!0.76$& $2.74\!\pm\!0.66$\\
2 & $7.57\!\pm\!1.06$& $5.51\!\pm\!1.30$& $6.15\!\pm\!1.35$& $4.56\!\pm\!1.00$\\
3 & $4.62\!\pm\!1.13$& $3.74\!\pm\!0.91$& $4.26\!\pm\!0.85$& $3.25\!\pm\!0.74$\\
4 & $3.63\!\pm\!0.65$& $3.31\!\pm\!0.75$& $3.35\!\pm\!0.74$& $2.74\!\pm\!0.73$\\
5 & $3.39\!\pm\!0.66$& $3.31\!\pm\!0.75$& $3.36\!\pm\!0.75$& $2.83\!\pm\!0.70$\\
6 & $1.87\!\pm\!0.59$& $2.09\!\pm\!0.73$& $2.05\!\pm\!0.73$& $1.82\!\pm\!0.71$\\
7 & $2.36\!\pm\!1.14$& $2.33\!\pm\!0.99$& $2.55\!\pm\!1.03$& $2.08\!\pm\!0.92$\\
8 & $3.54\!\pm\!1.78$& $3.23\!\pm\!1.44$& $2.51\!\pm\!1.33$& $2.21\!\pm\!1.15$\\
9 & $1.37\!\pm\!0.51$& $1.94\!\pm\!1.03$& $2.17\!\pm\!1.02$& $1.91\!\pm\!1.01$\\

\midrule
$\bm{\varnothing}$ & $\textbf{3.47}\!\bm{\pm}\!\textbf{0.37}$& $\textbf{3.09}\!\bm{\pm}\!\textbf{0.50}$& $\textbf{3.34}\!\bm{\pm}\!\textbf{0.52}$& $\textbf{2.68}\!\bm{\pm}\!\textbf{0.50}$\\
\bottomrule
\end{tabular}
\end{table}

\subsubsection{Qualitative Evaluation}

\begin{figure}[t]
\centerline{
\subfloat[]{\includegraphics[height=142px]{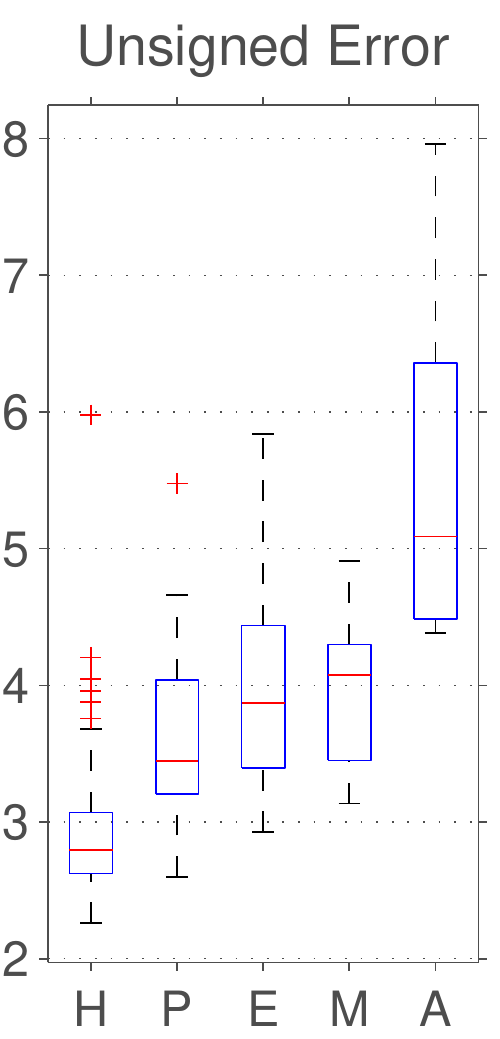}}
\hfill
\subfloat[]{\includegraphics[height=142px]{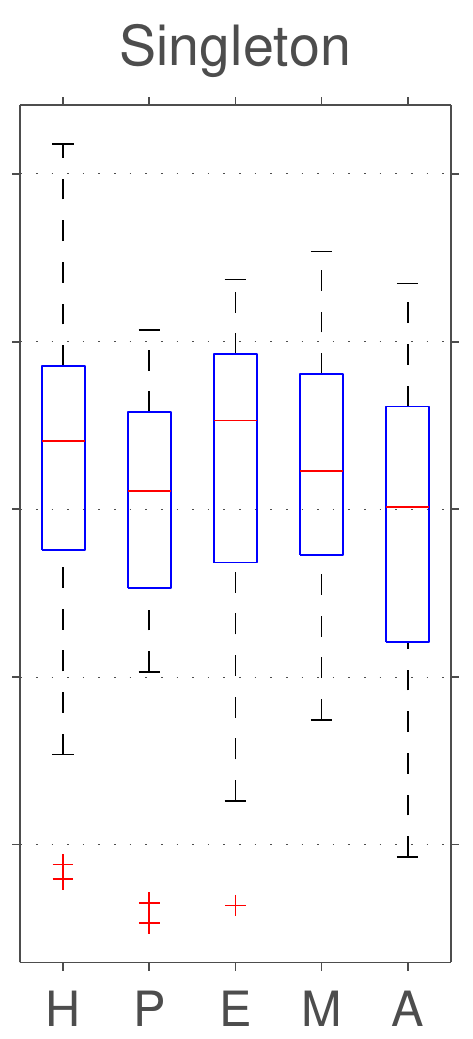}}
\hfill
\subfloat[]{\includegraphics[height=142px]{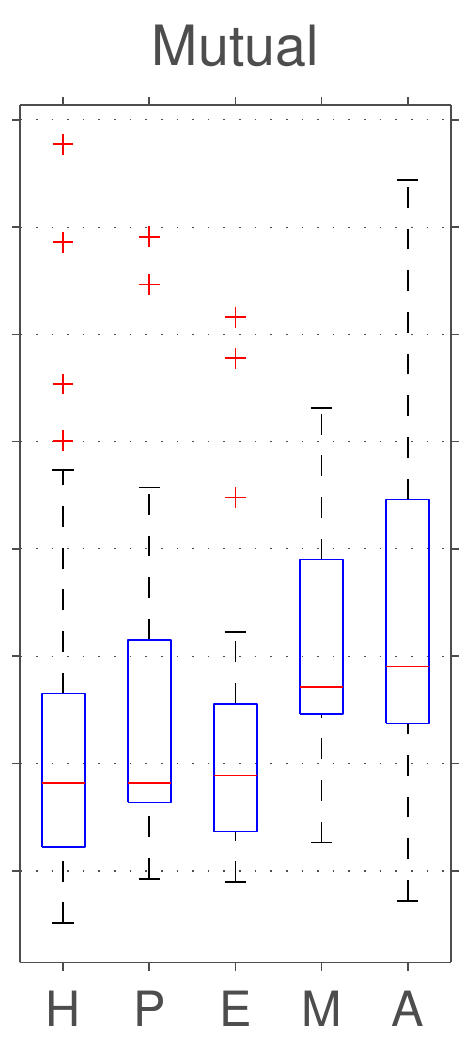}}
\hfill
\subfloat[]{\includegraphics[height=142px]{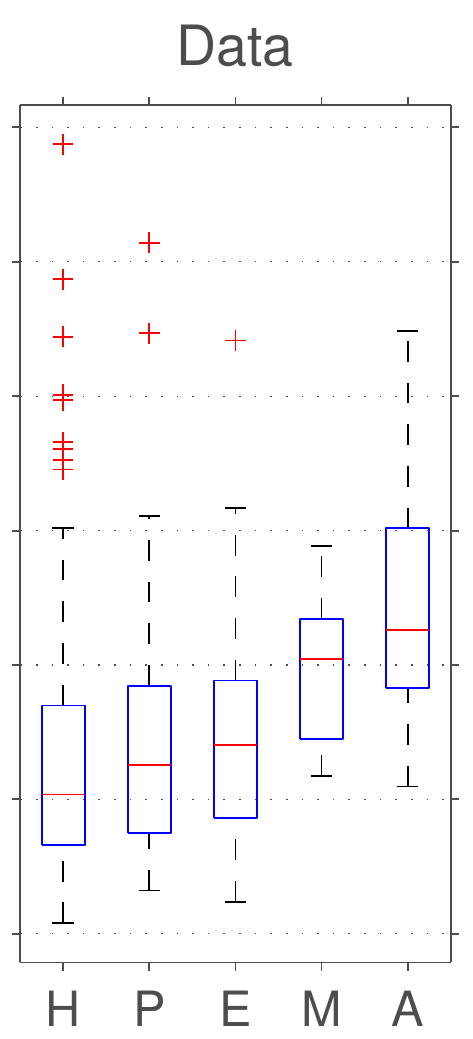}}
\hfill
\subfloat[]{\includegraphics[height=142px]{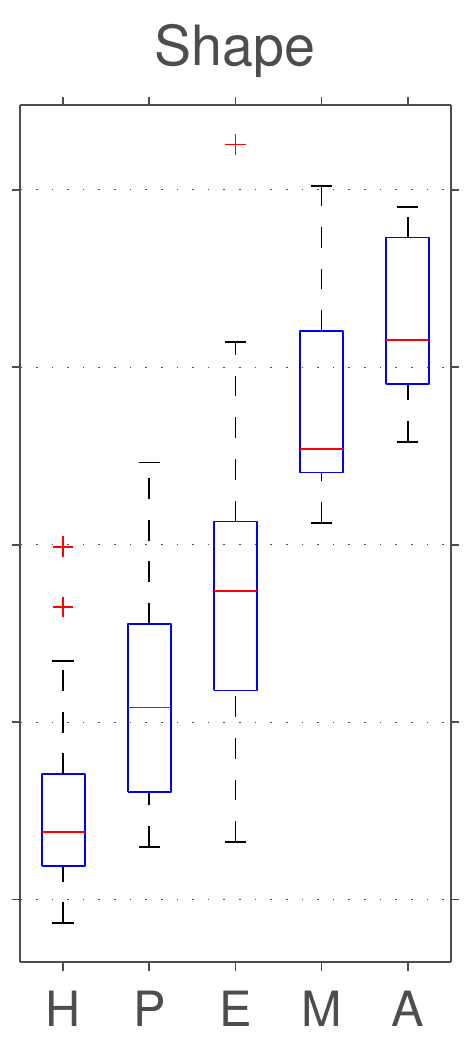}}
}
\caption{Compares different terms of the objective function (b-e) with the unsigned error~(a) for \textbf{h}ealthy as well as glaucomatous scans (\textbf{P}PG, PG\textbf{E}, PG\textbf{M} and PG\textbf{A}). While "Shape" is very discriminative for the glaucomatous scans, "Mutual" and "Data" correlate well with the unsigned error.}
\label{fig:eval-objective}
\end{figure}

A key property of our model is the \textit{inference of full probability distributions over segmentations} $q_c$ and $q_b$, instead of only modes thereof. This allows us to rate the quality of the prediction as a whole as well as indicate regions with low certainty, or classify a scan as normal or potentially pathological. To this end, we evaluated the different terms of the objective function \eqref{eq:J-functional}. Fig.~\ref{fig:eval-objective} reports average function values of four terms (b-e) and compares them to the unsigned error (a). Singleton entropy (b) and mutual information (c) are the two summands of the negative entropy of~$q_c$, given in~\eqref{eq:entropy-q-c}. The data~(d) and shape~(e) terms represent the first two summands of~\eqref{eq:J-functional}, introduced in Section~\ref{chap:p(c|y)} and~\ref{chap:p(c|b)}. 

The shape term, which measures how much the data-driven distribution~$q_c$ differs from the shape-driven expectation~$\mb{E}_{q_b}[\log p(c|b)]$, is highly discriminative between healthy and pathological scans. The mutual information on the other hand exhibit a good correlation with the unsigned error. It measures the dependence between variables $c_{k,j}$ and $c_{k-1,j}$. Imaging two variables $c_{k,j}$ and $c_{k-1,j}$ each having a single strong peak in $q_{c;k,j}$ and $q_{c;k-1,j}$. Their joint probability $\qcjointpure$ will show almost no dependency. On the other hand, if we have several possibilities for each variable caused e.g. by poor data terms, then their dependency increases and thereby the mutual information. We will use these two terms in the forthcoming evaluation.
\\[0.15cm]
\textbf{Classification.} A state-of-the-art method for the clinical diagnosis of glaucoma is based upon NFL thickness, averaged for example over the whole scan or one of its four quadrants 
\cite{bowd2001,leung2005,chang2009,leite2011}.
Estimates of the NFL thickness for all circular scans were obtained using the software of the Spectralis OCT device, version 5.6. We compared this established method against the second summand of the objective function \eqref{eq:J-functional}, as discussed above. Using the same setup as in \citet{bowd2001}, we report sensitivities for specificities of 70\% and 90\%, as well as the area under the curve (AUC) of the receiver operating characteristic (ROC)\footnote{The AUC can be interpreted as the probability, that a random pathological scan gets assigned a higher score than a random healthy scan.}, see Table \ref{tab:classification}. In all cases, our shape-based discriminator performs at least as good as the best thickness-based one.
Especially for pre-perimetric scans, which feature only subtle structural changes,
our approach improves diagnostic accuracies significantly. For this most interesting group, Fig.~\ref{fig:quality-index} (a) provides ROC curves of the two overall best performing NFL measures and our shape measure.
{\setlength{\tabcolsep}{0.1cm}
\begin{table}
\centering
\caption{Comparison of sensitivities for NFL-based features, measuring average thickness in different parts of the scan, and our global shape based feature. Bold numbers indicate the highest detection rate for the respective specificity and glaucoma class.}
\label{tab:classification}
\begin{tabular}{l c c c c c c c c c c c c c}
\toprule
Specificity &&\multicolumn{3}{c}{\textbf{70\%}} & & \multicolumn{3}{c}{\textbf{90\%}} & & \multicolumn{3}{c}{\textbf{AUC}} \\
Type &&\textit{PPG}& \textit{PGE}& \textit{PGM}& & \textit{PPG}& \textit{PGE}& \textit{PGM} && \textit{PPG}& \textit{PGE}& \textit{PGM} \\
\midrule
Average&& 68.2 & 90.9 & \textbf{100.0} && 36.4 & 86.4 & \textbf{100.0} && 0.72 & 0.93 & \textbf{1.00} \\
Superior&& 63.6 & 81.8 & 92.3 && 45.5 & 77.3 & 76.9 && 0.78 & 0.84 & 0.90 \\
Inferior&& 45.5 & 72.7 & 92.3 && 13.6 & 31.8 & 53.8 && 0.69 & 0.77 & 0.89 \\
Temporal&& 63.6 & \textbf{95.5} & \textbf{100.0} && 54.5 & 90.9 & \textbf{100.0} && 0.74 & \textbf{0.95} & 0.99 \\
Nasal&& 36.4 & 63.6 & 92.3 && 18.2 & 45.5 & 61.5 && 0.51 & 0.74 & 0.89 \\
\vspace{-10pt} \\
\hdashline[1pt/1pt]
\vspace{-9.5pt} \\
Shape&& \textbf{77.3}& \textbf{95.5}& \textbf{100.0}&& \textbf{63.6}& \textbf{95.5}& \textbf{100.0}&& \textbf{0.84} & \textbf{0.95} & \textbf{1.00}  \\
\bottomrule
\end{tabular}
\end{table}
\\[0.15cm]
\textbf{Global Quality.} We obtained a global quality measure, by combining the mutual information and the shape term. Given the values for all scans, we re-weighted both terms into the ranges $[0,1]$ and took their sum. Thereby we could establish a quality index that had a very good correlation of $0.82$ with the unsigned segmentation error. See Fig.~\ref{fig:quality-index} (b) for a plot of all quality index/error pairs and a linear fit thereof. The estimate of this fit and the true segmentation error differs on average by only $0.51\,\mu m$. This shows that the model is able to additionally deliver the quality of its segmentation.
\begin{figure}
\centering
\subfloat{\includegraphics[height=140px]{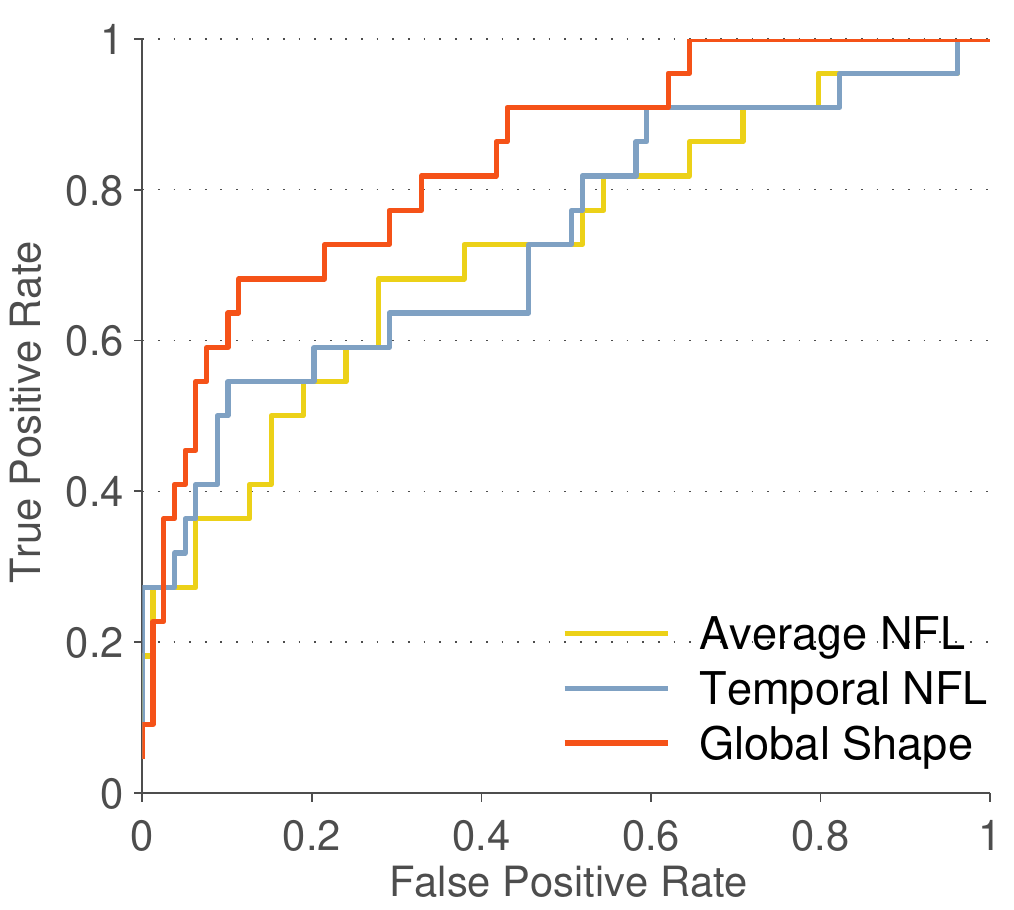}}
\hfill
\subfloat{\includegraphics[height=140px]{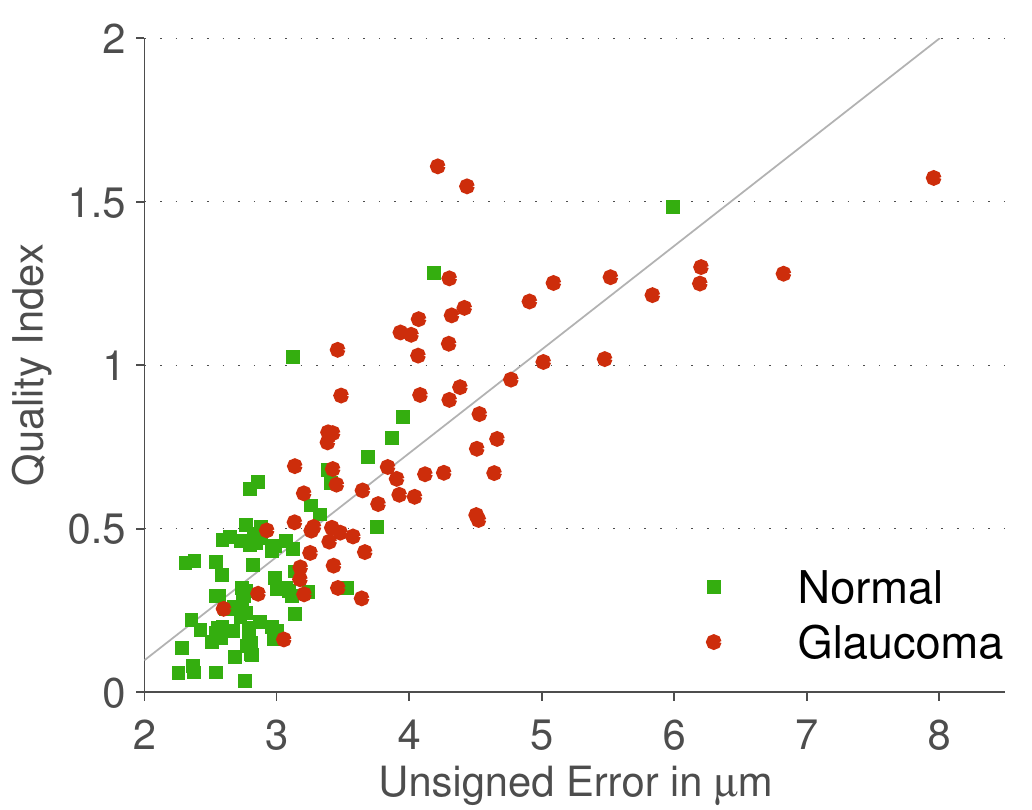}}
\caption{(left) ROC curves for the two overall best performing NFL-based classifiers and our shape prior based approach for pre-perimetric scans. (right) High correlation of our quality index, obtained by combining terms (c) and (e) from Fig.~\ref{fig:eval-objective}, with the actual unsigned error. On average, the estimated error (linear fit) differed by only $0.51 \,\mu m$ from the true segmentation error.}
\label{fig:quality-index}
\end{figure}
\\[0.15cm]
\textbf{Local Quality.} Finally, we determined a way to distinguish locally between regions of high and low model confidence. This could for example point out regions where a manual (or potentially automatic) correction is necessary. To this end we examined the local correlation (i.e. on a column-wise level) of the mutual information terms with the unsigned error. We calculated its mean for instances with segmentation errors smaller than 0.5 and bigger than 2 pixels. This yielded three ranges of confidence in the quality of the segmentation. For each image we fine-tuned these ranges by dividing by $\max(\mathrm{Quality\;Index(Current Image)},1)$.

Fig. \ref{fig:local-quality-glaukom} (a) shows a PGA-type scan with annotated segmentation, whose error is $6.83 \, \mu m$. The advanced thinning of the NFL and the partly blurred appearance caused the segmentation to fail in some parts of the scan. Close-ups (b) and (c) show that the model correctly identified those erroneously segmented regions. The average errors of the three categories are $4.67$, $5.43$ and $18.36\,\mu m$ respectively. Fig. \ref{fig:local-quality-healthy} (a), on the other hand, shows a scan from a healthy eye with a segmentation error of $2.83\,\mu m$, that is accompanied by a throughout positive quality rating.

We examined the accuracy of the local quality index numerically for all scans. Fig.~\ref{fig:local-quality-healthy} (b) reports the average unsigned error for normal (H) and glaucomatous scans (P-A) as well as all three grades of certainty, and compares it to the average segmentation error for each data set, given as black lines. As for the global quality index, also locally the model reflects very well the distinction between correct and erroneous regions. Fig.~\ref{fig:local-quality-healthy} (c) shows the ratio between the three quality ratings.

\begin{figure}[t]
\centerline{
\subfloat[]{\includegraphics[scale=0.5]{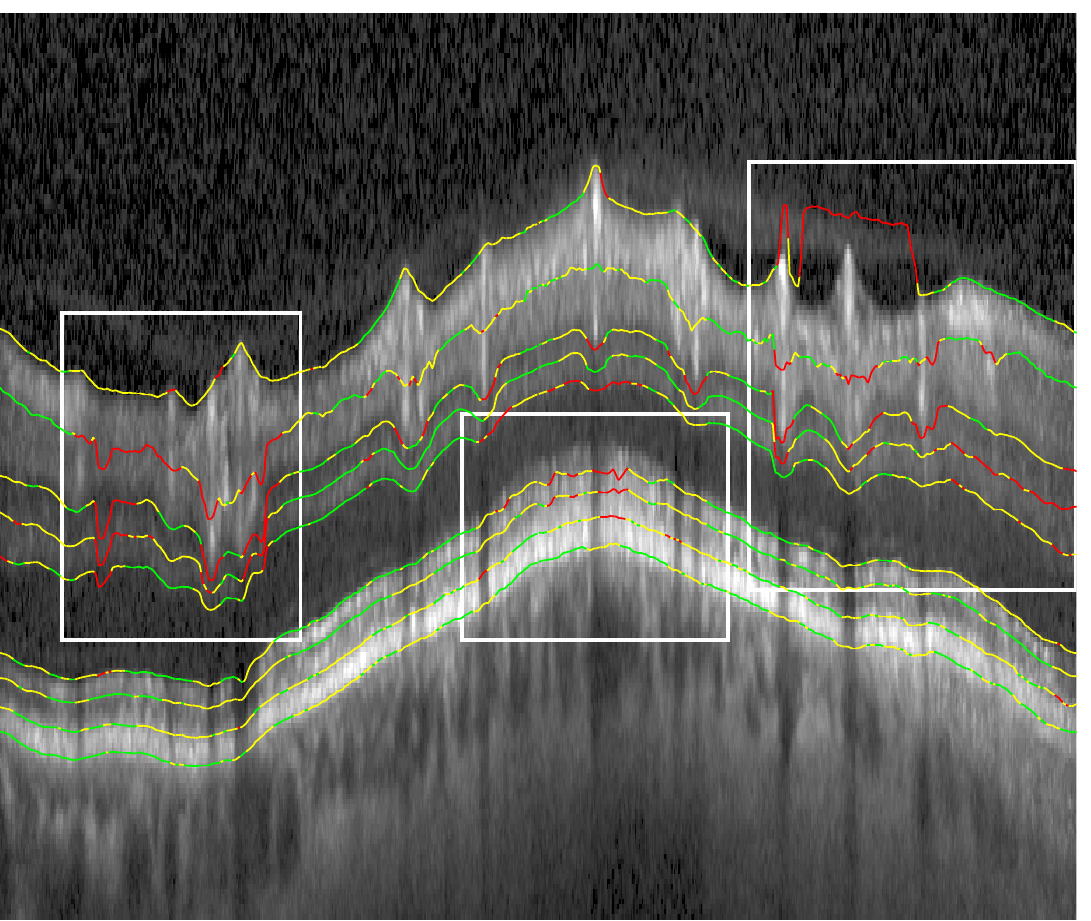}}
\hfill
\subfloat[]{
\includegraphics[scale=0.508]{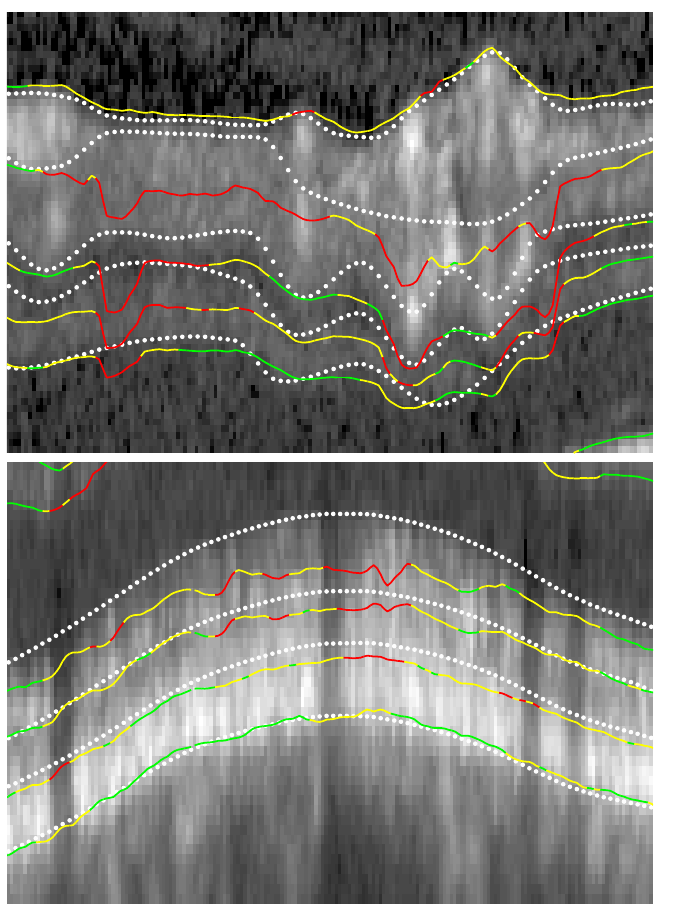}}
\hfill
\subfloat[]{\includegraphics[scale=0.5]{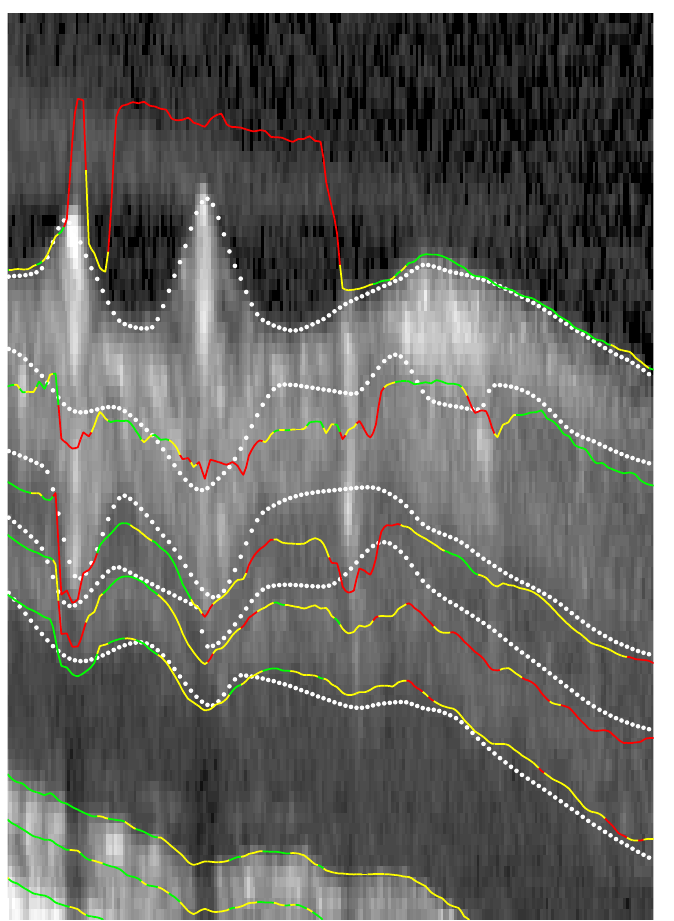}}
}
\caption{(a) An advanced primary open-angle glaucoma scan and the segmentation thereof ($E_\mathrm{unsgn} = 6.81\,\mu m$), augmented by the local quality estimates of the model, with red representing the most uncertain class. (b) and (c) Close-ups of the three areas, the model is (correctly) most insecure about. White dotted lines represent ground truth.}
\label{fig:local-quality-glaukom}
\end{figure}
\begin{figure}
\centerline{
\subfloat[]{\includegraphics[scale=0.44]{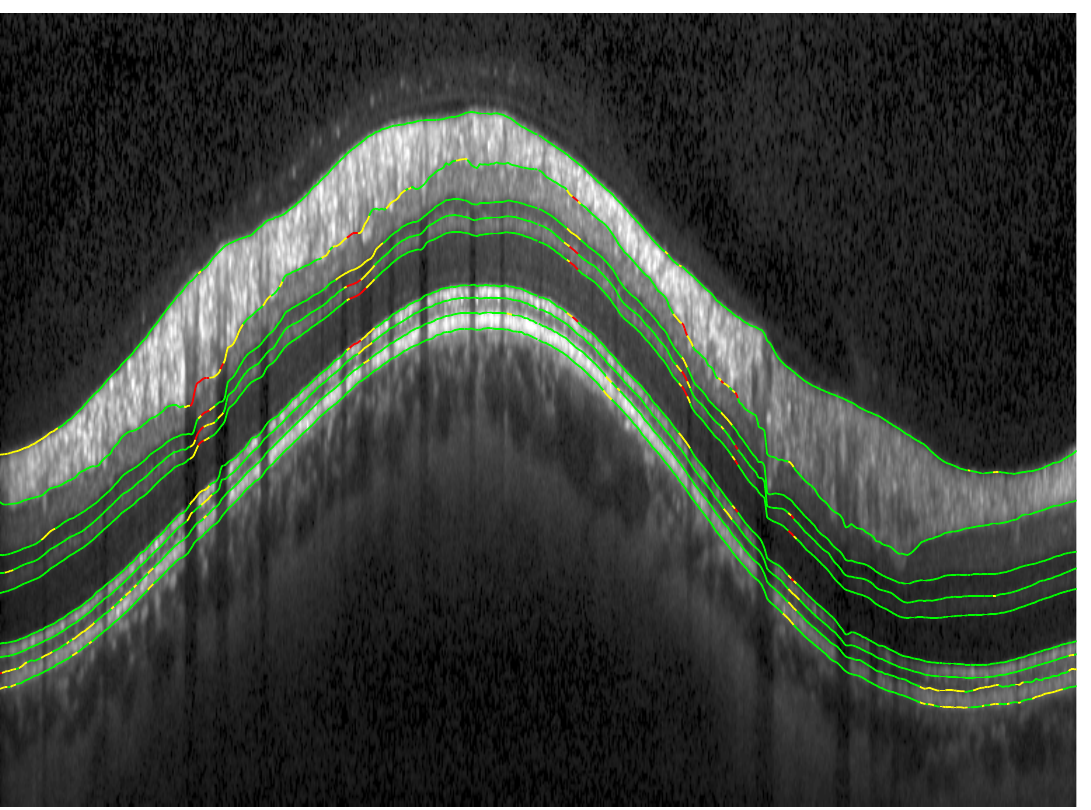}}
\hfill
\subfloat[]{\includegraphics[scale=0.44]{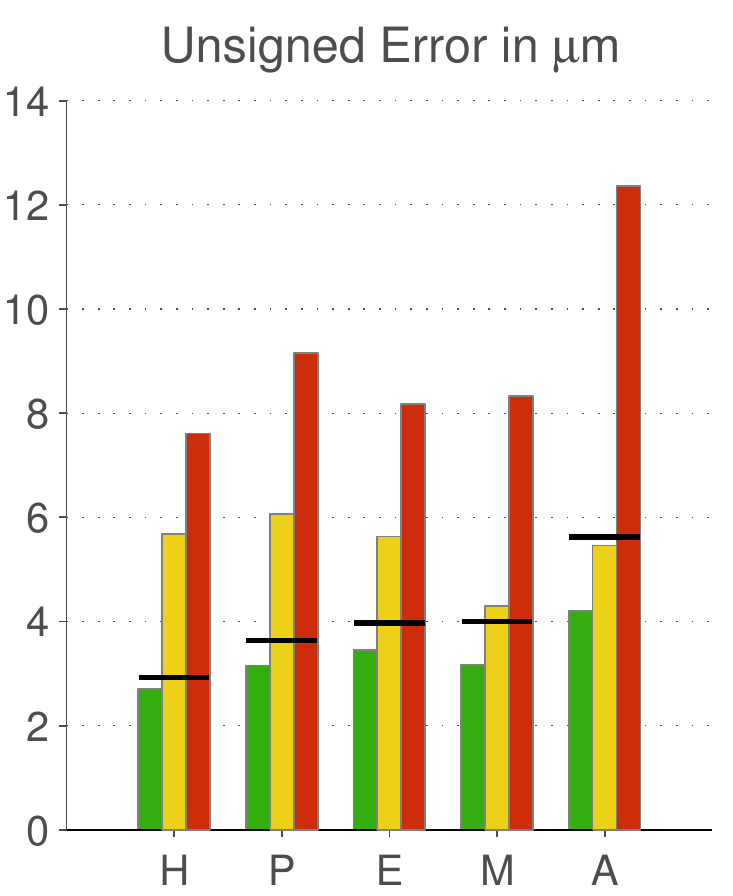}}
\hfill
\subfloat[]{\includegraphics[scale=0.44]{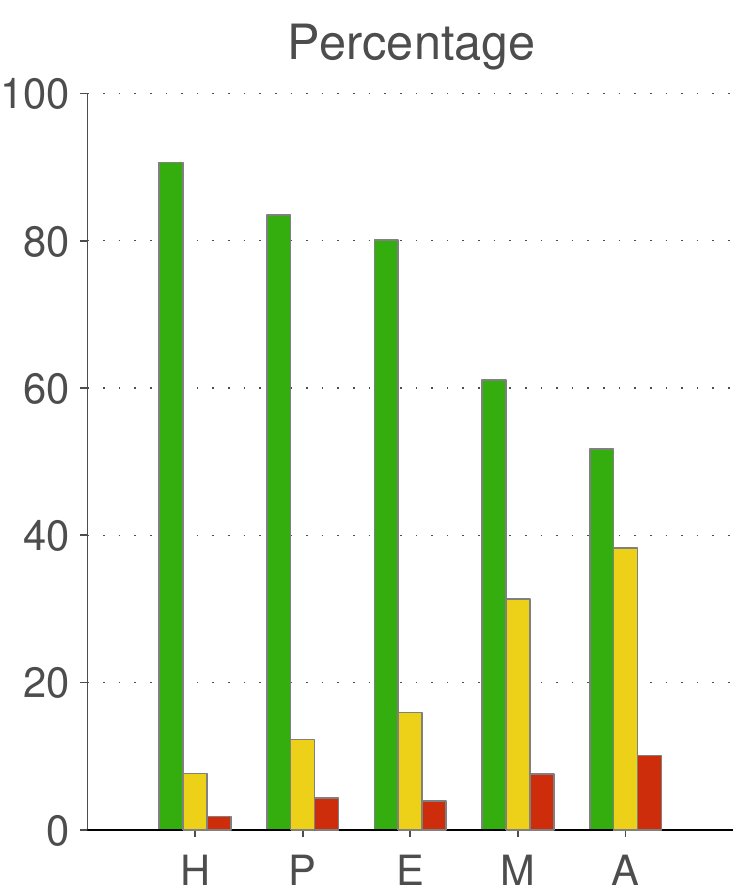}}
}
\caption{(a)~A healthy scan ($E_\mathrm{unsgn} = 3.05\,\mu m$)  accompanied by a very high model certainty. (b)~Average segmentation error for healthy as well as glaucomatous scans for each quality class. Black lines donate the average segmentation error for each data set (c.f.~Table~\ref{tab:performance-circ-scans}). (c)~Average partition into the three classes.}
\label{fig:local-quality-healthy}
\end{figure}

%% file: res_vol.tex
In contrast to 2-D scans, the labeling of OCT volumes is very time consuming, hence our data set only consisted of 35 samples. Thus we were left with less data points to train a shape model of much higher dimension. Consequently, we observed a reduced ability of $p(b)$ respectively $q_b(b)$ to generalize well to unseen scans. We tackled this problem by reducing the dimensionality of $p(b)$ and by interpolating it for intermediate columns, which fixed the problem only to some extent. 

We further pursued this idea and suppressed the connectivity between different B-scans, which corresponds to a block-diagonal covariance matrix $\Sigma$, where each block is obtained separately using PPCA. This significantly reduced the amount of parameters that had to be determined, and improved accuracy significantly. The last column in Table \ref{tab:performance-circ-scans} reports results for all boundaries.

\begin{figure}[t]
\centerline{
\subfloat{\includegraphics[width=0.25\textwidth]{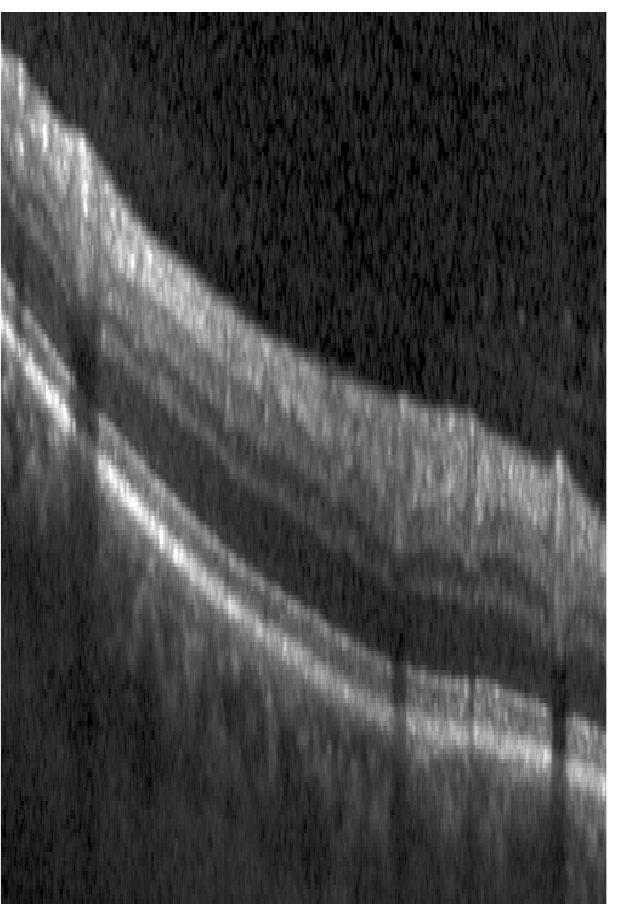}\hfill}
\subfloat{\includegraphics[width=0.25\textwidth]{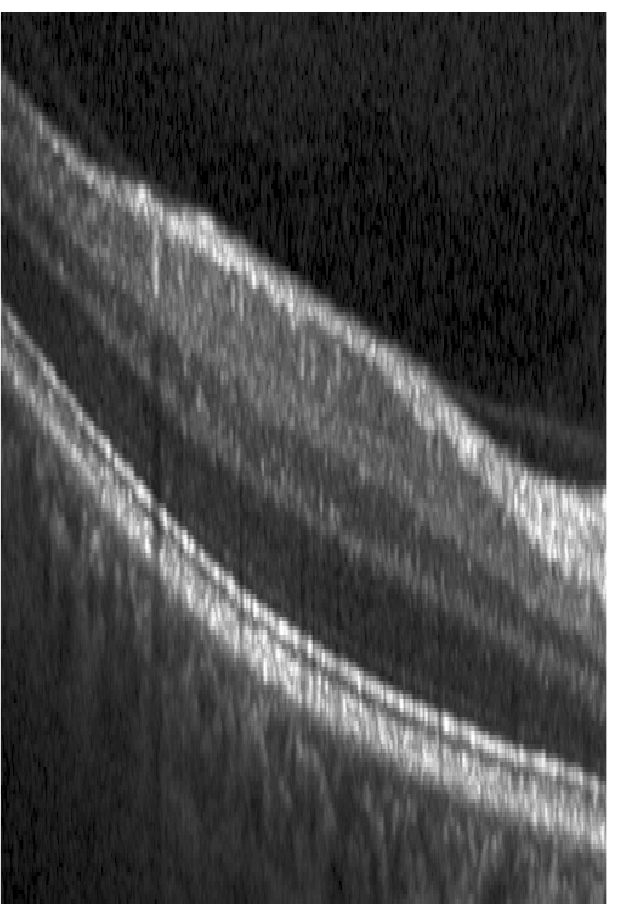}\hfill}
\subfloat{\includegraphics[width=0.25\textwidth]{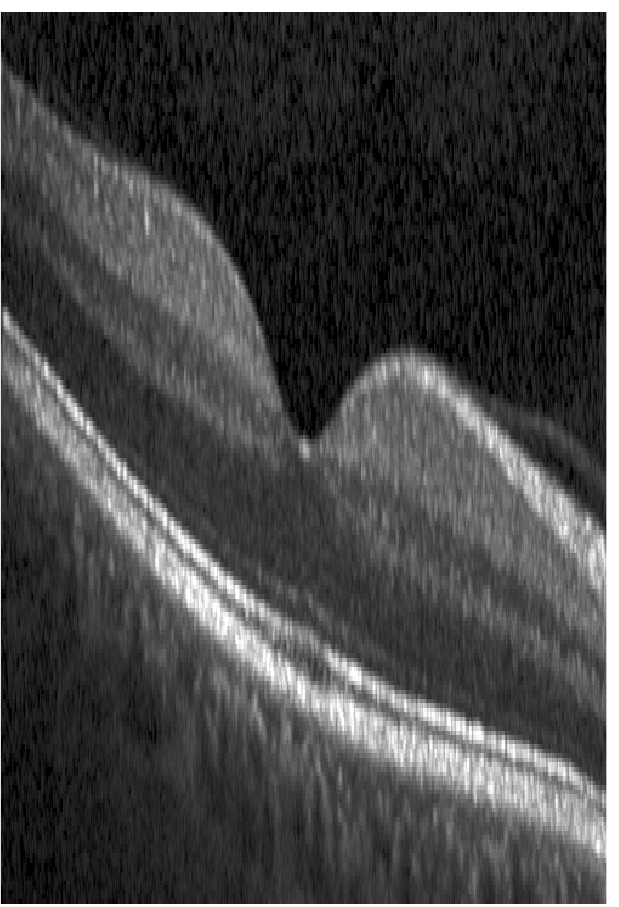}\hfill}
\subfloat{\includegraphics[width=0.25\textwidth]{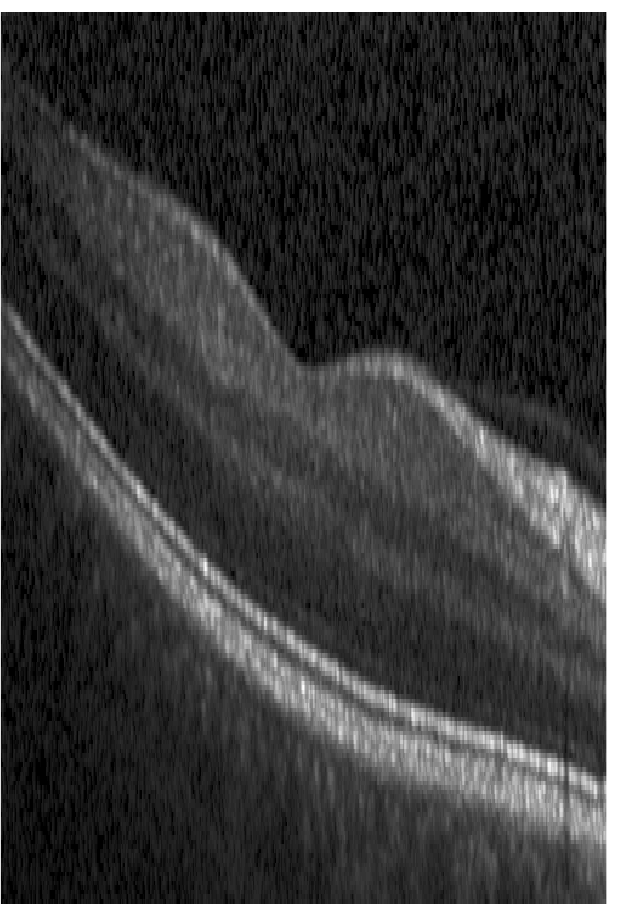}\hfill}
}
\vspace{-0.2cm}
\centerline{
\subfloat{\includegraphics[width=0.25\textwidth]{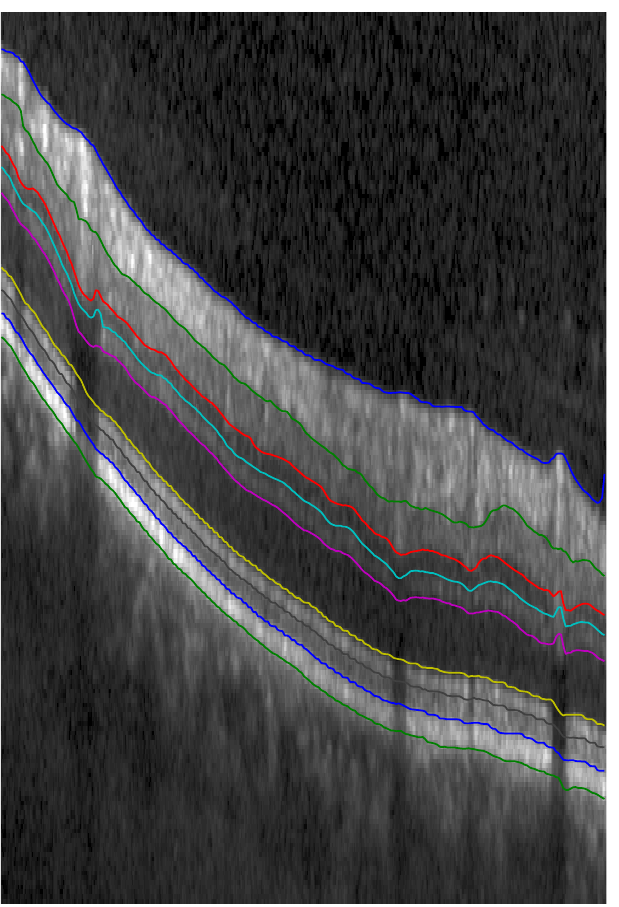}\hfill}
\subfloat{\includegraphics[width=0.25\textwidth]{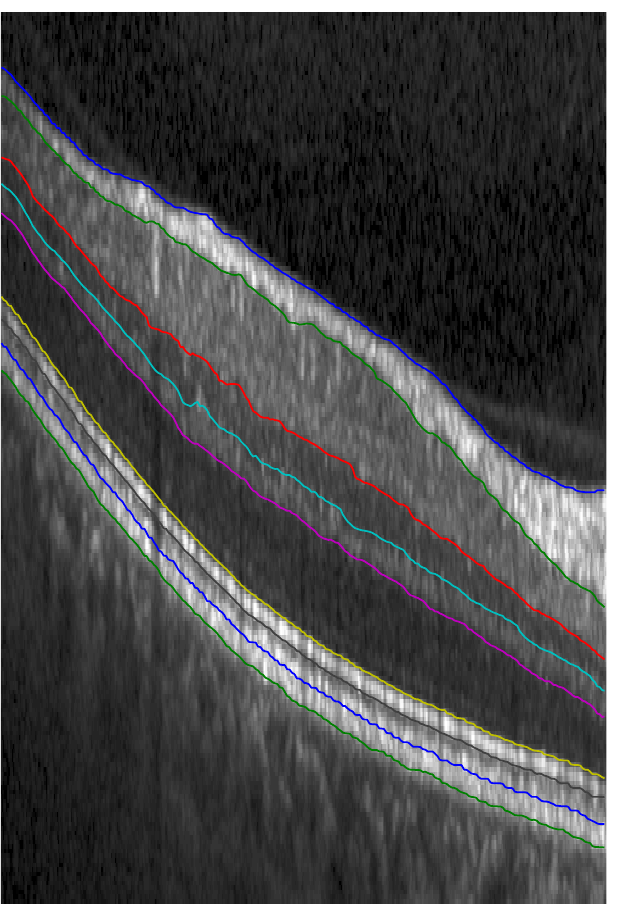}\hfill}
\subfloat{\includegraphics[width=0.25\textwidth]{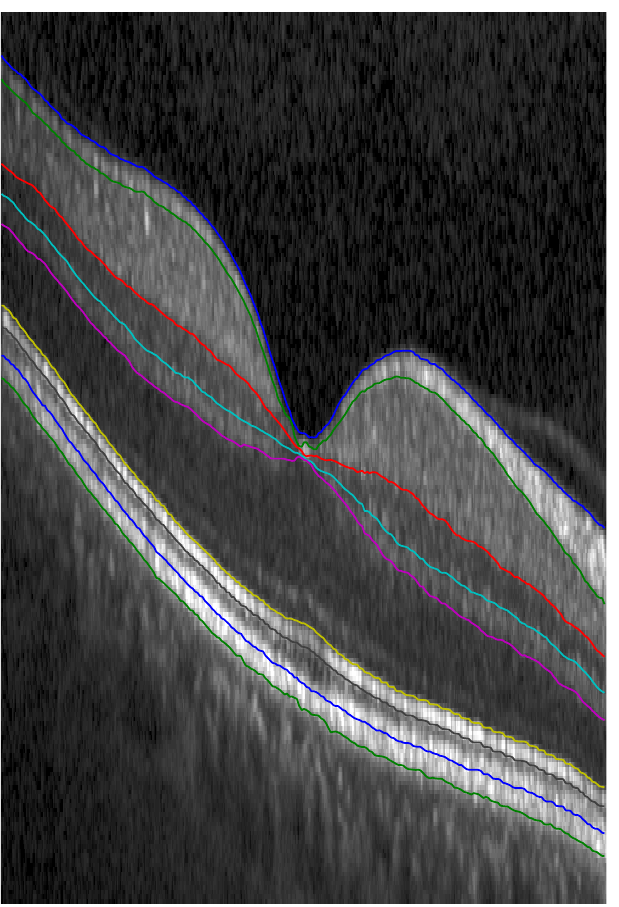}\hfill}
\subfloat{\includegraphics[width=0.25\textwidth]{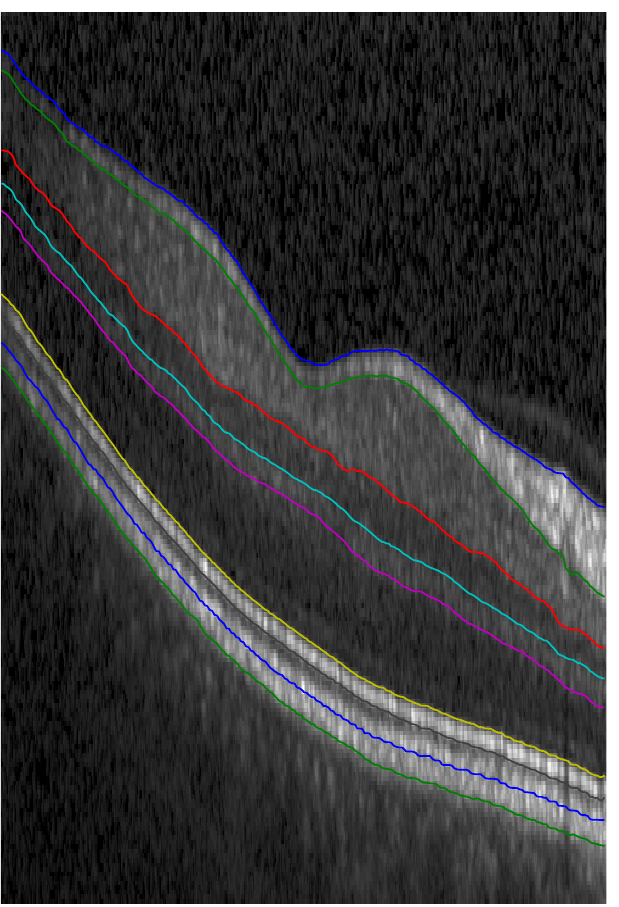}\hfill}
}
\caption{Four segmented B-Scans from regions 2, 6, 9 and 11 of the same volume ($E_\mathrm{unsgn} = 2.53\, \mu m$).}
\label{fig:vol-examples}
\end{figure}

The average segmentation error of $2.46\,\mu m$ is significantly smaller than for circular scans, as well as the standard deviation of $0.22\,\mu m$. Reasons are smoother boundary shapes and less severe texture artifacts caused by e.g. blood vessels. Representative for the average segmentation performance, Fig.~\ref{fig:vol-examples} shows B-scans of the same volume from four different regions, with an error of $2.53\,\mu m$ averaged over the all scans in the volume. 

%% file: res_disc.tex
A novel probabilistic approach for the segmentation of retina layers in OCT scans was presented. It incorporates \emph{global} shape information, which distinguishes it from most other approaches relying solely on local shape information. To obtain an approximate of the \emph{full} posterior distribution $p(c,b|y)$, we employed variational methods, which entail efficiently solvable optimization problems. 
We demonstrated the applicability of our approach for a variety of different OCT scans as well as the benefit of inferring full probability distributions over segmentations rather than segmentations as point estimates.

Especially for 3-D OCT volumes, our segmentation performance was significantly better than recently reported results from approaches that use no shape information \cite{vermeer2011,yang2010}, local hard-constrained shape information \cite{garvin2009}, local probabilistic shape information \cite{dufour2013,song2013} or sparse global shape information \cite{kajic2010}. Taking into account, for better comparability, only publications that used data sets obtained from the same OCT device as in this publication, the following trend evolves: 
\begin{itemize}
\item While \emph{no shape information} lead to only mediocre results: $6.20\,\mu m$ and $5.28\,\mu m$ for healthy and moderate glaucomatous data respectively \cite{vermeer2011}, 
\item adding \emph{local shape information via hard constraints} yielded improved segmentation performance: $3.54\pm0.56\,\mu m$ as evaluated by \citet{dufour2013} but comparable to the model proposed by \citet{garvin2009}. 
\item Additionally using \emph{probabilistic local constraints}, \citet{dufour2013} recently again boosted performance to $3.03\pm0.54\,\mu m$. 
\item Finally, by adding \emph{global shape information}, we could in turn improve segmentation performance to $2.46\pm0.22\,\mu m$. 
\end{itemize}
Although this clearly seems to support the use of global shape information for regularization, keep in mind that a concluding comparison can only be carried out using the same data set.
Nevertheless, we believe that these results highlight the usefulness of global shape regularization for the segmentation of retinal layers in OCT images.
Reported time requirements vary greatly, and our running time of \unit[60]{s} is slower than the \unit[18]{s} and \unit[15]{s} reported by \citet{dufour2013} and \citet{yang2010}, but faster than the remaining approaches cited above.

We also evaluated the performance for healthy and pathological 2-D circular scans and, in both cases, obtained good results. The only exception was the group of most advanced glaucomatous scans, which was caused mainly by the appearance models. Being trained on healthy data, the Gaussian distribution that models the NFL is not able to recognize instances with near-zero layer thickness. 
Therefore, a useful extension could be to define a mixture of Gaussians for each appearance class, adding patches centered below or above pixel~$(i,j)$, which model its surrounding but not the layer/boundary itself. Additionally, given more pathological examples especially for PGM and PGA, one could learn a pathological shape prior and let the model choose the more probable shape prior given the initialization. Future work will try to improve performance for these extreme cases and also test the approach on other pathologies like age-related macular degeneration, if training data gets available.

Finally, we investigated different ways to utilize the inferred distributions~$q_c$ and~$q_b$. Experiments showed, that the model is quite sensitive to abnormal shapes and thus can act as a detector of glaucoma, with a higher sensitivity than established methods solely based on NFL thickness. This could relate to recent findings, that glaucoma causes a thinning of \emph{all} inner retinal layers: NFL, GCL, IPL and (to a lesser extent) INL \cite{tan2008}.
To confirm these promising results, further studies with more patients enrolled will be needed.
Another benefit of our approach is the ability to estimate the quality of the segmentation, altogether for the whole scan or for each boundary position separately. In the context of screening large patient databases, the former could be a valuable tool to minimize the effort of the physician in reassessing the results.
The latter could facilitate a automatic or manual post-processing, targeted specifically at regions with a high error probability. A thorough investigation of these regions could reveal a suitable approach, and will be part of our future work. 

To facilitate further research in the area of OCT segmentation and related areas, we publish our source code together with documentation on our project page: \url{http://graphmod.iwr.uni-heidelberg.de/Project-Details.132.0.html}.

%% file: appendix.tex
\small
\section{Calculating $\Omega_{k,j}$ and $\omega_{1,j}$}
\label{app:expectation}

In Sec. \ref{chap:p(c|b)} we outlined the steps necessary to make explicit the expectation w.r.t $q_b$ for the terms of $\log p(b|c)$, represented by $(\omega_{1,j})_n$ and $(\Omega_{k,j})_{m,n}$. This section will derive both terms, starting with $(\omega_{1,j})_n$:
\begin{align*}
(\omega_{1,j})_n =&\; \mb{E}_{q_b}[\log p(b_{1,j} = n|b_{\sm{j}})] \\
=& \int_b  q_b(b) \log\mc{N}\big(b_{1,j}=n;(\mu_{j|\sm j})_1,(\Sigma_{j|\sm j})_{1,1}\big) \,  db \\
=& \; C - \frac{1}{2(E_{j|\sm{j}})_{1,1}}\Big(n^2 - 2n(\mb{E}_{q_b}[\mu_{j|\sm j})_1]+ \mb{E}_{q_b}[\big((\mu_{j|\sm j})_1\big)^2]\Big)\,.
\end{align*}
Using the definition of \eqref{eq:conditional-sec-summand} for $(\mu_{j|\sm j})_1$, abbreviating the $k$-th row of $\Sigma_{j | \sm j} K_{j,\sm{j}}$ with $\factor{k}$ and moving all terms independent of $\bar{\mu}$ to $C$
\begin{align}
\label{eq:expectation-p-c-b}
\nonumber (\omega_{1,j})_n =& \; C - \frac{1}{2(E_{j|\sm{j}})_{1,1}} \big(2(n-\mu_{1,j})\factor{1}\mb{E}_{q_b}[b_{\sm j}] +\factor{1}(\mb{E}_{q_b}[b_{\sm j}b_{\sm j}^T] - 2\mu_{\sm j}\mb{E}_{q_b}[b_{\sm j}])\factorT{1} \big) \,,\\
\intertext{and finally replacing the expectations with the respective moments given in \eqref{eq:moments} yields}
=& \; C - \frac{1}{2(E_{j|\sm{j}})_{1,1}}\big(2(n-\mu_{1,j})\factor{1}\bar{\mu}_{\sm{j}} +\factor{1}(\bar{\Sigma}_{\sm{j},\sm{j}} + \bar{\mu}_{\sm{j}}\bar{\mu}_{\sm{j}}^T - 2\mu_{\sm{j}}\bar{\mu}_{\sm{j}}^T)\factorT{1} \big) \,.
\end{align}

The terms $(\Omega_{k,j})_{m,n} = \mb{E}_{q_b}[\log p(c_{k,j} = n|c_{k-1,j} = m,b)]$ are the product of two Gaussians and therefore again Gaussian, modulo normalization. A lengthy derivation, using the formula for the product of two Gaussians, yields the same equation as above, with indices~$1$ replaced by~$k$ and a different constant $C$.

\section{Optimization of $q_b(b)$}
\label{app:optimization}
We showed in Sec. \ref{chap:optimization-q-b}, that the optimization of the objective function \eqref{eq:complete_functional} w.r.t. to the parameters of $q_b$ can be done in closed form. This section will detail terms $\tilde{p}$ and $\tilde{P}$ introduced there, which capture the dependencies of $\Omega_{k,j}$ and $\omega_{1,j}$ on the parameters of $q_b$.

\subsection{Derivation of $\tilde{P}$} 
Only considering terms in \eqref{eq:expectation-p-c-b} depending on $\bar{\Sigma}$, we obtain
\begin{equation*}
\label{eq:detail-omega-term}
(\omega_{1,j})_n(\bar{\Sigma}) = - \frac{1}{2(E_{j|\sm{j}})_{k,k}} \factor{k}\bar{\Sigma}_{\sm{j},\sm{j}}\factorT{k}
\end{equation*}
and similar for $(\Omega_{k,j})_{m,n}(\bar{\Sigma})$. Since $\factor{k}$ is of dimension $1 \times N_bM-N_b$ we introduce an expanded version $\factort{k} \in \mathbb{R}^{N_bM}$ padded with zero entries, such that $\factort{k}\Sigma\factortT{k} = \factor{k}\Sigma_{\sm{j},\sm{j}}\factorT{k}$. Note that $(\Omega_{k,j})_{m,n}(\bar{\Sigma})$ and $(\omega_{1,j})_n(\bar{\Sigma})$ are independent of $m$ and $n$ and thus have identical entries for all $(m,n)$, and therefore $\mathbb{E}_{p}[c] = c$. 
Using $b^TBb = \la bb^T,B \ra$, we obtain for~\eqref{eq:vectorized-second-term}
\begin{align*}
-\sum_{j=1}^M \Big( (q_{c;1,j})^T \omega_{1,j}(\bar{\Sigma}) + &\sum_{k=2}^{N_b} \big \langle (\qcjointpure)^T,\Omega_{k,j}(\bar{\Sigma}) \big\rangle \Big)\\
 &= \frac{1}{2} \Big(\sum_{j=1}^M \|q_{c;1,j}\|_1 \la \frac{1}{(E_{j|\sm{j}})_{1,1}}\factortT{1}\factort{1},\bar{\Sigma} \ra  \\
 & \qquad +\sum_{k=2}^{N_b} \|\qcjointpure\|_1 \la \frac{1}{(E_{j|\sm{j}})_{k,k}}\factortT{k}\factort{k},\bar{\Sigma} \ra\Big) \\
\label{eq:optimization_wrt_q_b_sigma_rewrite}&= \frac{1}{2}\big\la\sum_{j=1}^M \sum_{k=1}^{N_b} \tilde{P}_{k,j},\bar{\Sigma} \big\ra = \frac{1}{2} \la \tilde{P},\bar{\Sigma} \ra
\end{align*}

\subsection{Derivation of $\tilde{p}$} 
Again, we begin by singling out terms of \eqref{eq:expectation-p-c-b}, dependent on $\bar{\mu}$
\begin{equation*}
\label{eq:detail-omega-term2}
(\omega_{1,j})_n(\bar{\mu}) = - \frac{1}{2(E_{j|\sm{j}})_{1,1}}\big(2n\factor{1}\bar{\mu}_{\sm{j}} -2\mu_{1,j}\factor{1}\bar{\mu}_{\sm{j}} +\factor{1}(\bar{\mu}_{\sm{j}}\bar{\mu}_{\sm{j}}^T - 2\mu_{\sm{j}}\bar{\mu}_{\sm{j}}^T)\factorT{1} \big)
\end{equation*}
and accordingly for $(\Omega_{k,j})_{m,n}(\bar{\mu})$. The first term is dependent on $n$ and thereby on $q_c$, whereas the remaining terms are again independent on $n$ and $q_c$ sums out. Using $\factort{1}$ as introduced above, we plug $\omega_{1,j}(\bar{\mu})$ and $\Omega_{k,j}(\bar{\mu})$ into \eqref{eq:vectorized-second-term}
\begin{align*}
-\sum_{j=1}^M  &\Big( (q_{c;1,j})^T \omega_{1,j}(\bar{\mu}) +  \sum_{k=2}^{N_b} \big\langle (\qcjointpure)^T,\Omega_{k,j}(\bar{\mu}) \big\rangle \Big) \\
&= \frac{1}{2} \sum_{j=1}^M \sum_{k=1}^{N_b} \frac{1}{(E_{j|\sm{j}})_{k,k}} \Big(2\big(\mb{E}_{q_c}[c_{k,j}]-\mu_{k,j}\big)\factort{k}\bar{\mu} + \big\la \factortT{k}\factort{k},\bar{\mu}(\bar{\mu} - 2\mu)^T \big\ra \Big)\\
&= \frac{1}{2} \sum_{j=1}^M \sum_{k=1}^{N_b} 2\tilde{p}_{k,j}^T\bar{\mu} + \big\la \tilde{P}_{k,j},\bar{\mu}(\bar{\mu} - 2\mu)^T\big\ra \\
&=  \tilde{p}^T\bar{\mu} + \frac{1}{2} \la \tilde{P},\bar{\mu}(\bar{\mu} - 2\mu)^T \ra,
\end{align*}
where $\tilde{P}$ was defined in the previous section.
